%% file: acl_latex.tex
\documentclass[11pt]{article}

\usepackage[]{acl}
\usepackage{graphicx}
\usepackage{float}
\usepackage{subcaption}

\usepackage{adjustbox}
\usepackage{times}
\usepackage{latexsym}
\usepackage{utfsym}
\usepackage{pifont}
\usepackage{graphicx} 
\usepackage{caption}  
\usepackage{amssymb}
\usepackage{mathrsfs}
\usepackage[T1]{fontenc}

\usepackage[utf8]{inputenc}
\usepackage{amsmath}
\usepackage{cleveref}
\crefname{section}{§}{§§}
\Crefname{section}{§}{§§}

\usepackage{microtype}

\usepackage{inconsolata}

\usepackage{graphicx}
\usepackage{multirow}
\usepackage{makecell}
\usepackage{amsmath}
\usepackage{booktabs}
\definecolor{high}{HTML}{ff3333}  
\definecolor{low}{HTML}{77aef7}

\title{Think Twice, Click Once: Enhancing GUI Grounding via\\ Fast and Slow Systems}
\author{
\textbf{Fei Tang}\textsuperscript{1} \quad \quad
\textbf{Yongliang Shen}\textsuperscript{1} \quad \quad
\textbf{Hang Zhang}\textsuperscript{1} \\
\textbf{Siqi Chen}\textsuperscript{1} \quad \quad
\textbf{Guiyang Hou}\textsuperscript{1} \quad \quad
\textbf{Wenqi Zhang}\textsuperscript{1} \quad \quad
\textbf{Wenqiao Zhang}\textsuperscript{1} \\
\textbf{Kaitao Song}\textsuperscript{2} \quad \quad
\textbf{Weiming Lu}\textsuperscript{1} \quad \quad
\textbf{Yueting Zhuang}\textsuperscript{1} \\
\textsuperscript{1}Zhejiang University \\
\textsuperscript{2}Microsoft Research Asia \\
\{\texttt{flysugar}, \texttt{syl}\}@zju.edu.cn
}

\begin{document}
\maketitle
\begin{abstract}
Humans can flexibly switch between different modes of thinking based on task complexity: from rapid intuitive judgments to in-depth analytical understanding. However, current Graphical User Interface (GUI) grounding systems which locate interface elements based on natural language instructions rely solely on immediate prediction without reasoning, struggling to understand complex interface layouts with nested structures and hierarchical relationships, limiting their effectiveness on complex interfaces. Inspired by human dual-system cognition, we present \textsc{Focus}, a novel GUI grounding framework that combines fast prediction with systematic analysis. The framework dynamically switches between rapid and deliberate processing through an adaptive system switching based on task complexity, optimizing both efficiency and accuracy. 
\textsc{Focus} decomposes grounding into progressive stages: interface summarization, visual focused analysis, and precise coordinate prediction. This structured decomposition enables systematic understanding of both interface layouts and visual relationships. Extensive experiments show that \textsc{Focus} achieves state-of-the-art performance using only 300K of the training data with a 2B parameter model compared to existing approaches. 
\textsc{{Focus}} demonstrates superior performance particularly in complex GUI scenarios, achieving 77.4\% average accuracy on ScreenSpot and 13.3\% on the more challenging ScreenSpot-Pro. Our analysis reveals the effectiveness of this dual-system approach while demonstrating its potential for improving complex GUI interaction scenarios\footnote{https://github.com/sugarandgugu/Focus}.
\end{abstract}

\section{Introduction}
GUI interactions are fundamental to modern human-computer interaction, driving the development of GUI agents that can automate complex interface operations \cite{he2024pcagentsleepai,hu2024dawnguiagentpreliminary}. A critical capability for these agents is GUI grounding: the ability to accurately locate and interpret interface elements based on natural language instructions. This foundational skill directly impacts an agent's effectiveness in understanding interface semantics and executing operations across diverse scenarios \cite{cheng2024seeclickharnessingguigrounding,yang2024ariauivisualgroundinggui}.

Early GUI agents primarily relied on structured information such as XML or DOM trees for element localization \cite{wang2024mobileagentautonomousmultimodalmobile, zhang2023appagentmultimodalagentssmartphone,li2024appagentv2advancedagent}. While this approach provided reliable structural information, it faced significant limitations in practical applications due to information redundancy and accessibility issues \cite{gou2024navigatingdigitalworldhumans,cheng2024seeclickharnessingguigrounding}.
Recent advances in multimodal large language models (MLLMs) \cite{bai2023qwenvlversatilevisionlanguagemodel,lu2024deepseekvlrealworldvisionlanguageunderstanding,ye2024mplugowlmodularizationempowerslarge,chen2024internvlscalingvisionfoundation,liu2023visualinstructiontuning} have enabled direct screenshot-based approaches for GUI grounding. For instance, SeeClick \cite{cheng2024seeclickharnessingguigrounding} leverages visual GUI pretraining to enhance element localization, while ShowUI \cite{lin2024showuivisionlanguageactionmodelgui} incorporates advanced visual-language processing techniques.
\begin{figure}[t!] 
    \centering
    \includegraphics[width=0.49\textwidth]{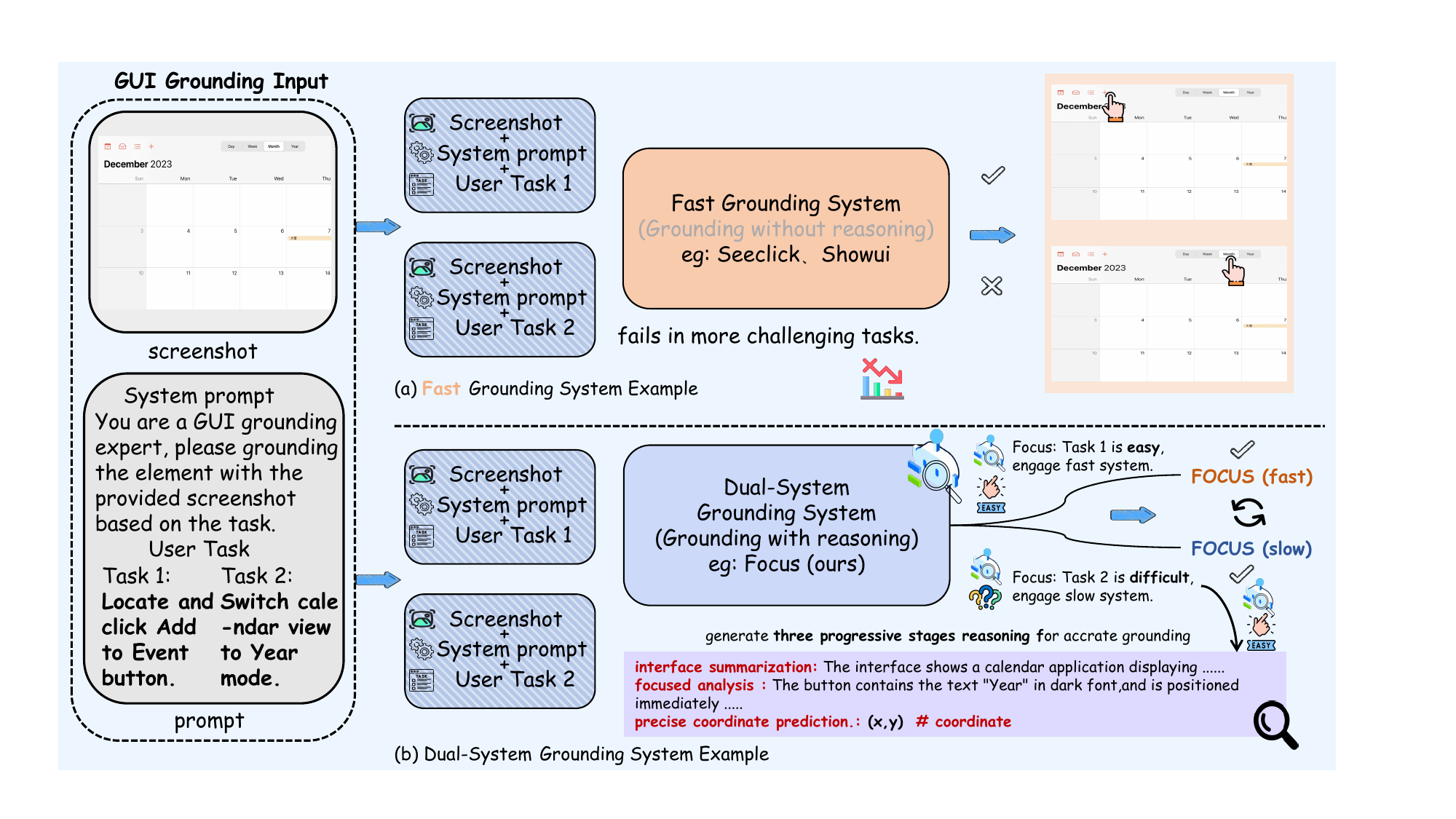} 
    \caption{Comparison of GUI grounding approaches.
    (a) Fast grounding system (\textit{e.g.}, SeeClick \cite{cheng2024seeclickharnessingguigrounding}, ShowUI \cite{lin2024showuivisionlanguageactionmodelgui}) directly predicts target locations without explicit reasoning. (b) Our \textsc{Focus} framework introduces a dual-system approach combining fast grounding with deliberate analysis, dynamically switching between systems based on task complexity.} 
    \label{fig:motivation} 
\end{figure}

However, current MLLM-based GUI grounding methods \cite{cheng2024seeclickharnessingguigrounding, lin2024showuivisionlanguageactionmodelgui,yang2024ariauivisualgroundinggui} predominantly follow a direct prediction paradigm, attempting to locate target elements immediately from screenshots without deeper analysis.
This approach, while efficient, faces two critical limitations: (1) \textit{Limited Interface Understanding}: These methods struggle with complex interfaces containing multiple windows, nested menus, and hierarchical structures, lacking systematic analysis of interface layouts and element relationships. (2) \textit{Insufficient Visual Analysis}: While accurate element identification requires understanding multiple visual attributes (\textit{e.g.}, color, shape, position) and their contextual relationships,  current approaches attempt to directly predict target locations like human's fast system, without engaging the slow system's deliberate analysis that is characteristic of human analytical thinking. 

The efficient grounding and deeper analysis reflect a fundamental characteristic of human cognition in task processing. Humans employ two distinct cognitive systems when interpreting visual interfaces: a rapid, intuitive system for simple tasks and a deliberate, analytical system for complex scenarios \cite{kahneman2011thinking,evans2008dual}. 
Inspired by this, we propose \textsc{Focus}, a GUI grounding framework that combines fast grounding with deliberate analysis. As shown in Figure \ref{fig:motivation}, \textsc{Focus} preserves the efficiency of direct prediction while introducing a systematic analysis process that first comprehends the interface layout before localizing target elements. The framework dynamically switches between two system based on task complexity, enabling efficient handling of both simple and complex GUI interactions.

To train \textsc{Focus}, we generate structured training data that decomposes GUI grounding into three progressive stages: (1) interface summarization to capture overall layout structure, (2) focused analysis of relevant interface regions and their visual characteristics, and (3) precise coordinate prediction. This structured decomposition allows \textsc{Focus} to build comprehensive understanding from global context to local details. Through training on this multi-stage reasoning process, \textsc{Focus} develops robust generalization capabilities across diverse GUI scenarios.

Our main contributions are as follows:
\begin{itemize}
\item We propose \textsc{Focus}, a novel framework combining fast grounding with deliberate analysis for robust GUI grounding, drawing inspiration from human cognitive dual-process theory.
\item We introduce a hierarchical training approach that decomposes GUI grounding into progressive stages, enabling systematic interface understanding and visual analysis.
\item We develop an adaptive switching mechanism between fast and deliberate processing based on task complexity, optimizing both efficiency and accuracy.
\item Extensive experiments demonstrate \textsc{Focus} achieves state-of-the-art performance, with 1.4\% and 17.7\% improvements on ScreenSpot and ScreenSpot-Pro benchmarks respectively.
\end{itemize}
\section{Method}
\textsc{{Focus}} enhances GUI grounding through dynamically adapting its processing strategy between fast and slow systems based on task complexity. For simple tasks, it employs fast grounding system for efficient processing. When facing complex scenarios, it activates slow grounding system, which generates an a task-oriented \textbf{interface summarization} and conducts \textbf{focused analysis} guided by the task instruction. To train such an adaptive dual-system, as shown in Figure \ref{fig:overview}, we propose a data synthesis pipeline that decomposes GUI interactions into three progressive stages, constructing a 300K-sample dataset that captures human-like cognitive processes.

Our approach includes three main components: (1) Dual-System Data Synthesis (\cref{sec:3-1}) that constructs training data by decomposing GUI grounding into progressive stages, mimicking human's transition from rapid intuitive judgments to in-depth analytical thinking; (2) Dual-System FOCUS Training (\cref{sec:3-2}) that develops specialized model capabilities for both fast grounding and systematic analytical processing; (3) Adaptive System Switching (\cref{sec:3-3}) that dynamically transitions between fast and slow systems based on task complexity.
We detail our framework in the following sections.

\subsection{Dual-System Data Synthesis}
\label{sec:3-1}
To effectively train our dual-system model, we propose a progressive data synthesis strategy that systematically constructs training examples with increasing complexity while naturally differentiating between fast and slow cognitive processes. Our approach not only synthesizes training data but also adaptively separates examples based on their inherent complexity levels:
We begin with fast grounding data synthesis, employing ShowUI \cite{lin2024showuivisionlanguageactionmodelgui} for direct coordinate prediction. Successfully predicted cases form our fast grounding data, capturing straightforward tasks that can be solved through immediate perception. When initial predictions fail, the second stage enhances the process by incorporating \textbf{interface summarization} before prediction. The successful cases in this stage contribute to our slow grounding data, representing scenarios that require structural understanding. For the most challenging cases where both attempts fail, our final stage introduces \textbf{focused analysis} of specific element characteristics (location, shape, color) within the generated interface context. The complete reasoning chain is recorded as slow grounding data, capturing tasks that require detailed visual analysis. This progressive data synthesis strategy enables \textsc{Focus} to build comprehensive understanding from global context to local details.

\input{tables/dataset_count}
\subsubsection{Data Collection}
\begin{figure*}[t!]
\centering
\includegraphics[width=1.0\textwidth]{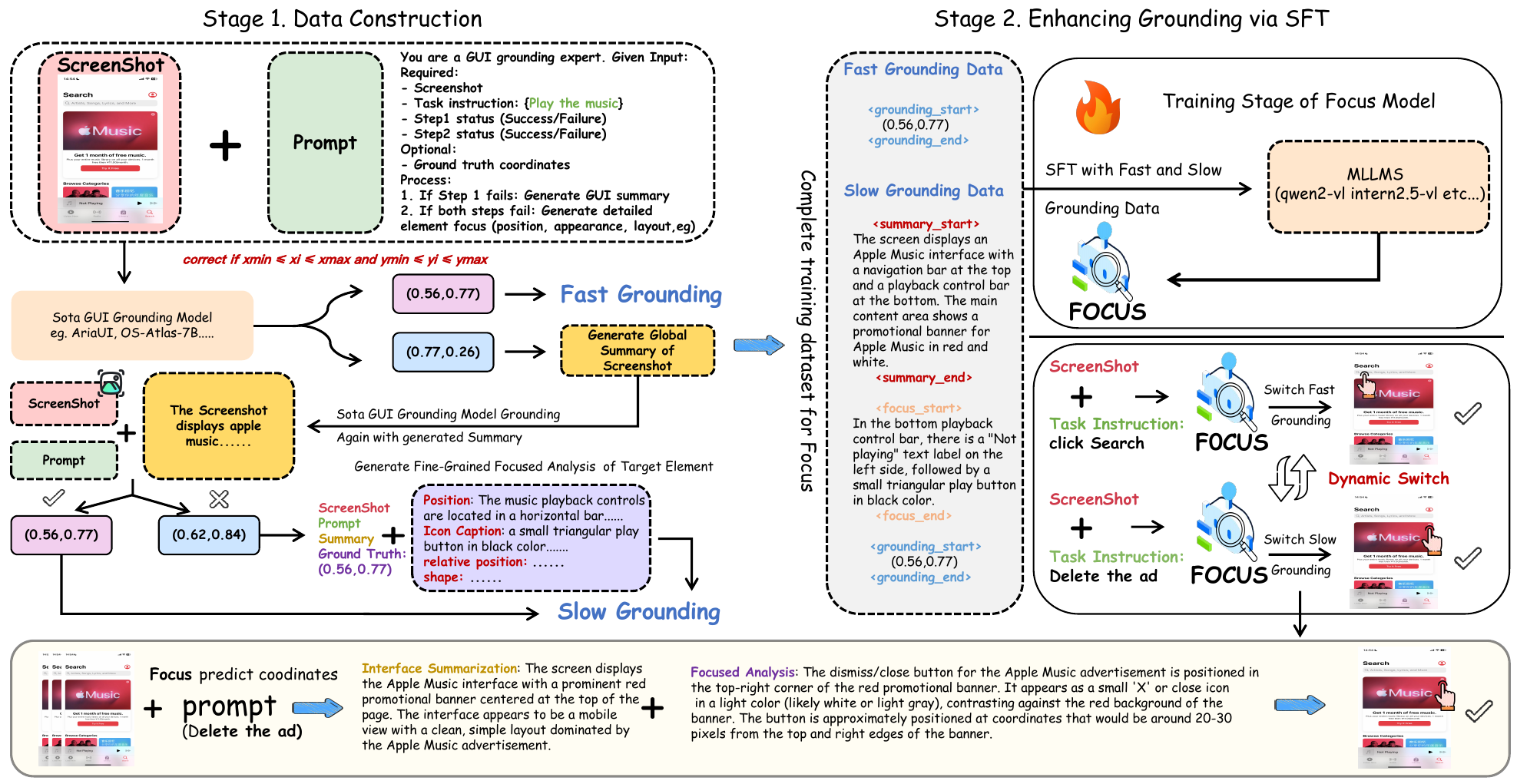}
\caption{Overview of \textsc{Focus} data construction and training process: the system performs interface summarization and task-oriented visual focused analysis for grounding. The middle shows complete examples of fast and slow grounding data. \textsc{Focus} dynamically switches between fast and slow grounding systems, with a complete example of slow-system grounding shown at the bottom.}
\label{fig:overview}
\end{figure*}

The GUI Grounding task requires the GUI grounding model to predict coordinates $c$ of the target element most relevant to the given screenshot $s$ and task instruction $i$. To train our dual-system model, we collected and processed GUI interaction data spanning web interfaces, mobile applications, and desktop software. Our final dataset contains 300K samples with diverse interaction scenarios, as detailed in Table \ref{tab:dataset_count}. Each data sample is structured as a triplet containing a task instruction, bounding box coordinates, and a corresponding screenshot. Following previous works \cite{cheng2024seeclickharnessingguigrounding, lin2024showuivisionlanguageactionmodelgui}, we normalize the target element's bounding box to center point coordinates $(x,y)$ in the $[0,1]$ interval. For example, as shown in Stage 2 of Figure \ref{fig:overview}, given the instruction "click the search button", the model outputs normalized coordinates like $(0.49, 0.33)$.
\subsubsection{Fast Grounding System}
Fast grounding system is similar to a direct cognitive simulation of human intuitive judgments in simple tasks, where the model immediately predicts target coordinates from the input (\textit{e.g.}, Seeclick \cite{cheng2024seeclickharnessingguigrounding}).
To synthesize fast grounding data, We employ ShowUI \cite{lin2024showuivisionlanguageactionmodelgui} to simulate human's rapid decision-making process. Fast grounding data synthesis process consists of two key process that simulate human intuitive judgments in simple tasks:
\begin{itemize}
    \item The model receives screenshot-instruction pairs, analogous to how humans quickly parse visual and textual information.
    \item Predicted coordinates $(x, y)$ are evaluated against ground truth using a perceptual matching criterion.
\end{itemize}

The criteria for determining the correctness of the prediction results are:
\begin{equation}
x_{\min} \leq x \leq x_{\max} \quad \text{and} \quad y_{\min} \leq y \leq y_{\max}
\label{eq1}
\end{equation}

Where $x,y$ represent the target coordinates predicted by the model, and $(x_{\min},y_{\min},x_{\max},y_{\max})$ denote the true coordinates of the target element's bounding box.

When predicted coordinates successfully fall within the bounding box, we classify the sample as fast grounding data. This classification reflects human-like rapid and intuitive task resolution. When fast grounding fails, it triggers a cognitive shift to a more deliberative, analytical mode (our slow grounding system) to handle complex interface scenarios.
\subsubsection{Slow Grounding System}
When the fast grounding system fails to accurately predict target element locations, we introduce a slow grounding system that follows a progressive process inspired by human analytical thinking for complex tasks:
\begin{itemize}
    \item \textbf{Interface Summarization}: The model first generates task-oriented summary of the GUI interface layout and structure based on the given instructions. This summary captures both the global interface organization and the hierarchical relationships between interface elements. We combine this generated summary with the original input for a second coordinate prediction using the GUI Grounding model. If the prediction satisfies the coordinate accuracy criteria defined in Equation \ref{eq1}, we classify it as Slow Grounding data.
    \item \textbf{Focused Analysis}: When summary-based prediction fails, we prompt the model conducts fine-grained visual focused analysis of potential target elements. This stage involves examining multiple visual attributes including relative positions, absolute locations, shapes, colors, and contextual relationships with surrounding interactive elements. This detailed analysis helps narrow down the search space and identify distinguishing features of the target element.
     \item \textbf{Precise Coordinate Prediction}: Finally, we integrate information from both previous stages, combining interface summarization for global context and focused analysis for local details as input for the final coordinate prediction. This progressive refinement from global understanding to precise localization mirrors human cognitive processes in solving complex tasks. The complete reasoning chain is recorded as Slow Grounding data.
\end{itemize}
This three-stage approach enables systematic analysis of complex GUI grounding tasks through progressive decomposition, enhancing model robustness across diverse interface scenarios.
\subsection{Dual-System \textsc{Focus} Training}
\label{sec:3-2}
\subsubsection{Training Protocol Modeling}
\textsc{Focus} simulate two cognitive systems to model the fast and slow systems of human cognition in GUI grounding. The fast grounding system is modeled to emulate human's rapid pattern recognition and immediate response capabilities, operating on direct visual-semantic mappings without intermediate analysis. The slow grounding system mirrors human's deliberate analytical process through three key components: interface layout comprehension, focused visual analysis, and reasoned decision making. Through our data construction pipeline, we systematically built a training dataset that captures these cognitive patterns, with 145K samples (48.3\%) representing fast cognitive processing and 150K samples (51.7\%) embodying slow analytical reasoning. Each slow grounding sample encodes the complete cognitive sequence - from initial interface comprehension to focused analysis and final decision making, enabling \textsc{Focus} to develop human-like cognitive flexibility in GUI interactions.
\subsubsection{Training Strategy}
To effectively train our dual-system model, we design specialized token structures that guide the model's reasoning process. These tokens serve as explicit markers, enabling the model to develop distinct capabilities for fast grounding and slow analytical processing.

We introduce three sets of special tokens to structure different reasoning chains:
\begin{itemize}
    \item Fast grounding tokens \textit{<|grounding\_start|>}, \textit{<|grounding\_end|>} encapsulate direct coordinate predictions, such as \textit{<|grounding\_start|>}$(0.46,0.78)$\textit{<|grounding\_end|>}.
    \item Interface summarization tokens \textit{<|summary\_start|>}, \textit{<|summary\_end|>} mark global layout analysis in the slow reasoning chain.
    \item Focused analysis tokens \textit{<|focus\_start|>}, \textit{<|focus\_end|>} designate detailed element analysis, followed by coordinate predictions.
\end{itemize}
Through these specialized tokens, the model develops distinct cognitive processes for both fast and slow thinking paths. This structured design not only ensures consistent modeling of cognitive behavior but also provides explicit guidance for different reasoning strategies.
\subsection{Adaptive System Switching}
\label{sec:3-3}
Effective switching between processing systems is crucial for optimizing both computational efficiency and analytical accuracy in GUI grounding tasks. To achieve this balance, we propose an adaptive systems switching that dynamically transitions between fast and slow systems based on token probabilities. 

During inference, for each input pair of GUI screenshot \( s \) and task instruction  \( i \), we denote the probability of generating the first token \( t \) as \(p(t \mid s, i)\). To control the switching process between systems, we introduce a scaling factor \( \alpha \) to adjust the activation threshold:
\begin{equation}
p_{\text{slow}}(t \mid s, i) = \alpha \cdot p(t = t_s \mid s, i)
\end{equation}
\begin{equation}
p_{\text{fast}}(t \mid s, i) = (1-\alpha) \cdot p(t = t_g \mid s, i)
\end{equation}
where \(t_s\) and \(t_g\) represent the probabilities of generating \textit{<|summary\_start|>} and \textit{<|grounding\_start|>} tokens, respectively. The system automatically activates the mode with the higher probability between \( p_{\text{slow}} \) and \( p_{\text{fast}} \).

Through token-based probability switching, \textsc{Focus} achieves optimal balance between computational efficiency and deep analysis capabilities.
\subsection{Implementation Details}
We built \textsc{Focus} on Qwen2-VL-2B-Instruct \cite{wang2024qwen2vlenhancingvisionlanguagemodels} and fine-tuned it for GUI grounding tasks. The model was fully fine-tuned on our constructed dataset of 300K fast and slow grounding examples for 3 epochs, with the visual backbone and MLP projections frozen. For optimization, we employed AdamW with a cosine annealing learning rate scheduler, setting the initial learning rate to 1e-4. and trained on 4 A100 (40GB) GPUs for 24 hours. We provide the inference details in Appendix \ref{sec:inference-details}. Additionally, an ablation study of $ \alpha $ is provided in Section \ref{sec:alpha_exp}, which explores the impact of different values of $ \alpha $ on system performance.
\section{Experiment}
\subsection{Experiment Setup}
To comprehensively evaluate \textsc{Focus}'s grounding capabilities, we conduct extensive experiments on two GUI grounding benchmarks: ScreenSpot \cite{cheng2024seeclickharnessingguigrounding} and ScreenSpot-Pro \cite{li2024screenspot-pro}. ScreenSpot contains 1,272 samples across mobile, desktop and web platforms, focusing on common interface scenarios and element types. Due to its limitations in evaluating professional software environments, ScreenSpot-Pro was introduced with 23 professional applications featuring high-resolution interfaces and complex layouts.

We compare \textsc{Focus} against both closed-source multimodal models (\textit{e.g.}, GPT-4V, Gemini-1.5-pro \cite{geminiteam2024gemini15unlockingmultimodal}) and open-source GUI grounding models (\textit{e.g.}, ShowUI \cite{lin2024showuivisionlanguageactionmodelgui}, CogAgent \cite{hong2024cogagentvisuallanguagemodel}). Following previous works, a prediction is considered correct if the predicted coordinates $(x, y)$ fall within the target element's bounding box. We report the average accuracy across all test cases as the evaluation metric.

\input{tables/screenspot}
\subsection{GUI Grounding Evaluation}
The evaluation results on ScreenSpot \cite{cheng2024seeclickharnessingguigrounding} are shown in Table \ref{tab:screenspot}. Among models with 2B parameters, \textsc{Focus} achieves the best performance with an average accuracy of 77.4\%, surpassing both ShowUI (75.1\%) and InfiGUIAgent (76.3\%). Notably, general-purpose multimodal models like GPT-4V and Gemini-1.5-pro achieve lower accuracy (16.7\% and 53.2\% respectively). We observe that \textsc{Focus} performs particularly well in challenging scenarios, achieving 78.2\% accuracy in icon/widget grounding on mobile platforms.
On the more challenging ScreenSpot-Pro \cite{li2024screenspot-pro} benchmark, \textsc{Focus} achieves state-of-the-art performance with 13.3\% overall accuracy, surpassing AriaUI (11.3\%) and CogAgent (7.7\%) despite having fewer parameters. Notably, \textsc{Focus} demonstrates strong icon/widget recognition capability, achieving 3.9\% accuracy compared to ShowUI's 2.6\%. 
The performance of ScreenSpot and ScreenSpot-Pro highlights the effectiveness of \textsc{Focus}'s dual-system approach, where the fast and slow systems dynamically switch based on task complexity, ensuring both efficient and accuracy.
\begin{figure*}[t!]
    \centering
    \begin{subfigure}[t]{0.48\textwidth}
        \centering
        \includegraphics[width=\textwidth]{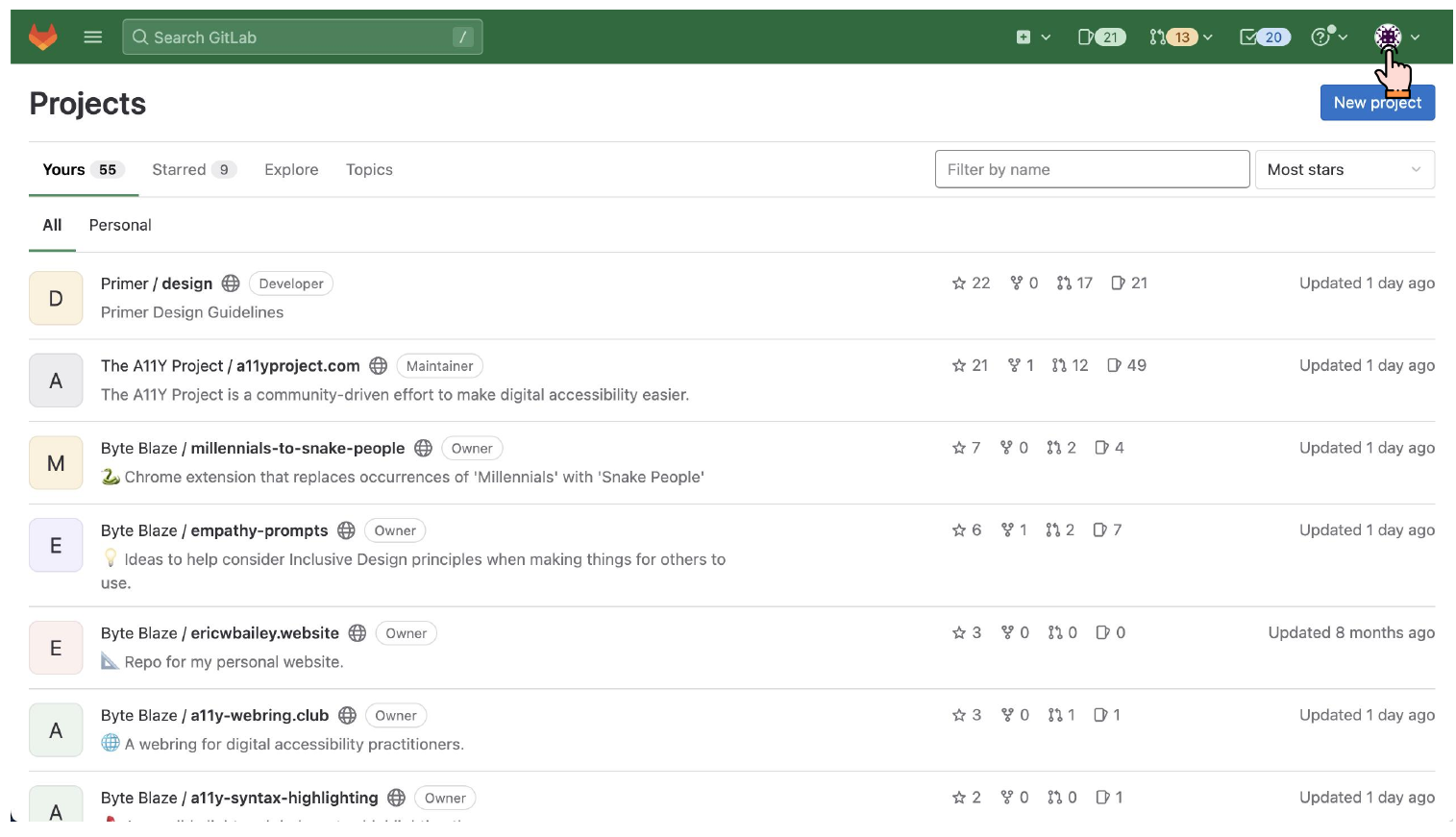}
        \caption{\ding{52} \textsc{\textbf{Focus}} achieves accurate grounding by generating interface summary and visual focusing analysis.}
        \label{fig:focus_case}
    \end{subfigure}
    \hfill
    \begin{subfigure}[t]{0.48\textwidth}
        \centering
        \includegraphics[width=\textwidth]{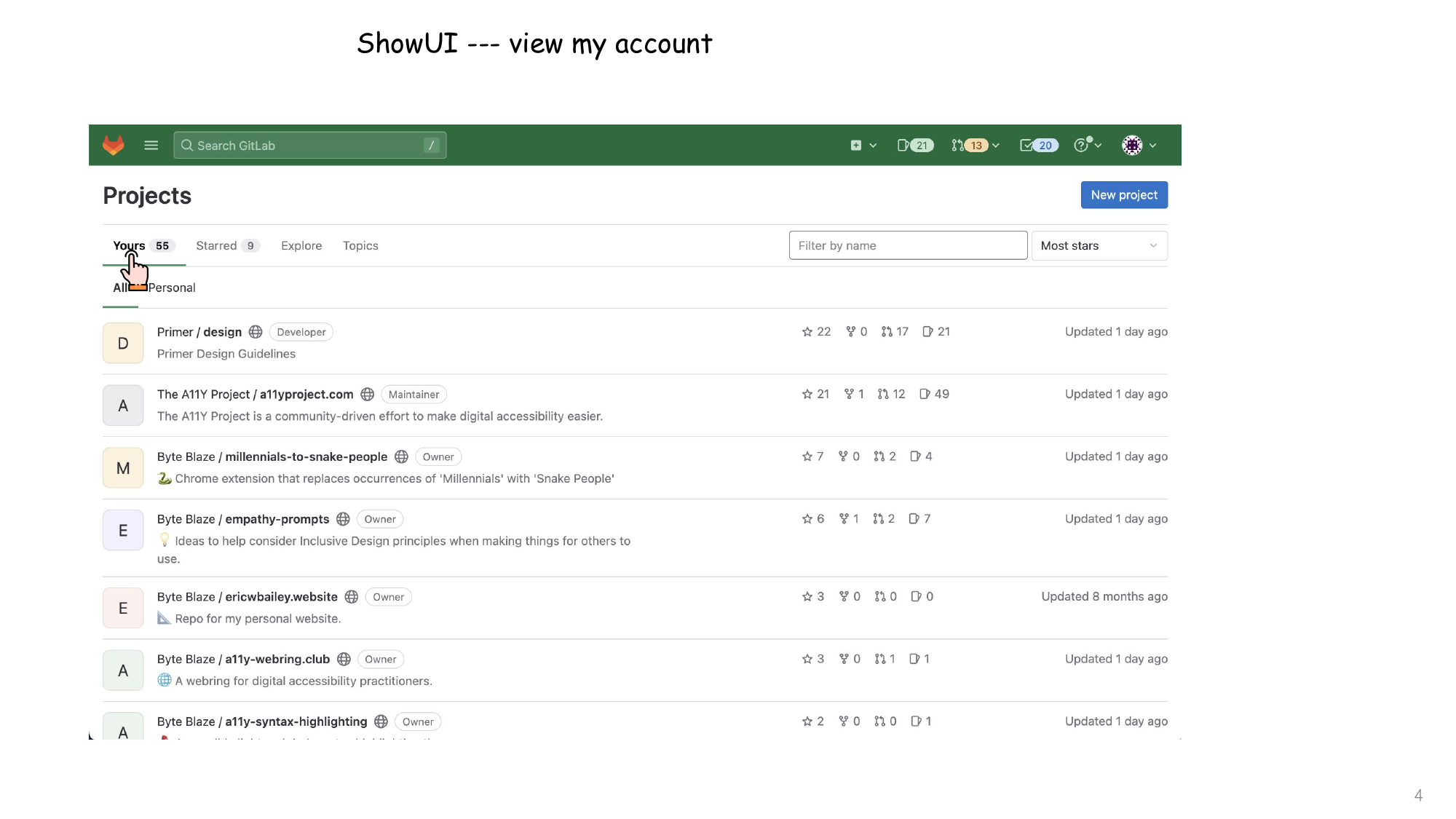}
        \caption{\ding{55} \textbf{ShowUI} \cite{lin2024showuivisionlanguageactionmodelgui} made an incorrect prediction in this scenario.}
        \label{fig:showui_case}
    \end{subfigure}
    \caption{Case study comparison between ShowUI and \textsc{Focus}. ShowUI struggles with complex GUI scenarios, while \textsc{Focus} excels through its dual-system.}
    \label{fig:case_study}
\end{figure*}
\subsection{Analysis}  
\input{tables/screenspot-pro}
\input{tables/ablation_study}
\subsubsection{Impact of Progressive Stage}
To validate the effectiveness of \textsc{Focus}'s dual-system architecture, we conducted ablation studies by creating three variants: removing interface summary generation (w/o Summary), removing visual focusing analysis (w/o Focus), and removing both components (w/o Both) from the slow system. As shown in Table \ref{tab:ablation}, removing both components leads to a significant performance drop from 77.4\% to 71.4\%. Individual component analysis shows that excluding interface summary and visual focusing causes 2.6\% and 4.2\% accuracy decreases respectively.
The performance impact is particularly evident in icon/widget localization tasks, as noted in SeeClick \cite{cheng2024seeclickharnessingguigrounding}. For instance, on mobile platforms, removing the visual focusing component reduces icon/widget localization accuracy from 78.2\% to 74.9\%. These results demonstrate that both components of \textsc{{Focus}}'s slow system are complementary and essential for robust GUI element grounding.

Our ablation studies suggest the importance of progressive stages in GUI grounding. When the focused analysis stage is removed (w/o Focus), though the model can capture high-level interface layouts, it may lack the granular visual reasoning needed for distinguishing subtle differences between interface elements. This limitation could be particularly problematic when dealing with elements that share similar visual attributes but serve different functions. Without the summarization stage (w/o Summary), the model attempts fine-grained visual analysis without first establishing contextual understanding, potentially leading to a semantic gap between local visual features and their roles in the broader interface hierarchy. When both stages are removed (w/o Both), the model essentially performs direct localization similar to the fast grounding system, which may be insufficient for complex interfaces with nested structures and intricate element relationships. These findings indicate that the progressive refinement from global understanding to detailed analysis could be key to robust GUI grounding, especially in sophisticated interface scenarios, where each stage contributes complementary strengths for element disambiguation.
\begin{figure}[t]
    \centering
    \includegraphics[width=1.0\linewidth]{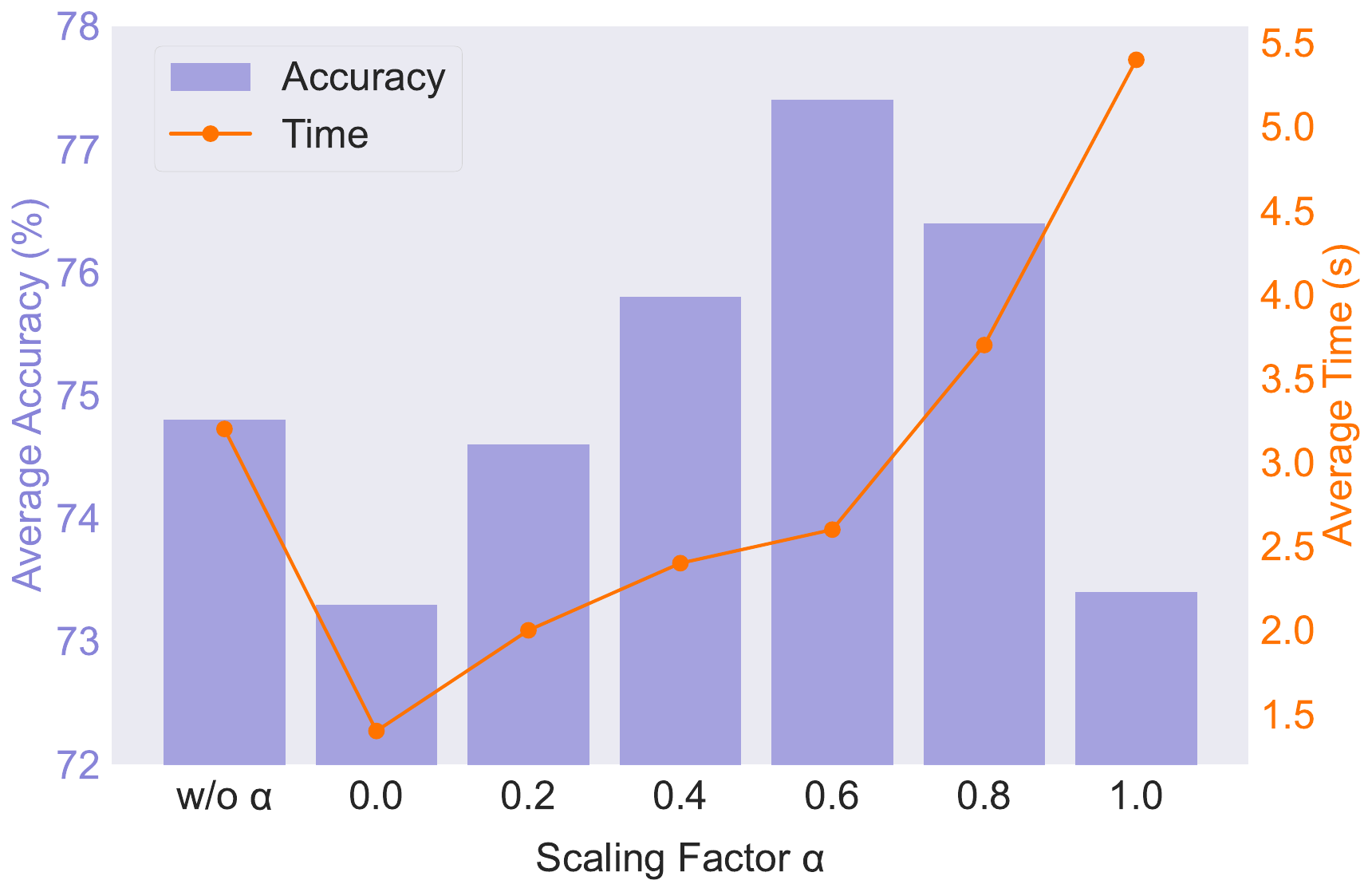}
    \caption{Impact of scaling factor $\alpha$ on \textsc{Focus}'s performance, where "w/o $\alpha$" represents baseline without Adaptive System Switching. At $\alpha$ = 0.6, \textsc{Focus} achieves +2.6\% accuracy improvement while reducing processing time by 0.6s compared to baseline, demonstrating effective balance between accuracy and efficiency.}
    \label{fig:alpha_exp}
\end{figure}
\subsubsection{Balancing Fast and Slow Systems} 
\begin{figure}[H] 
    \centering
    \includegraphics[width=0.49\textwidth]{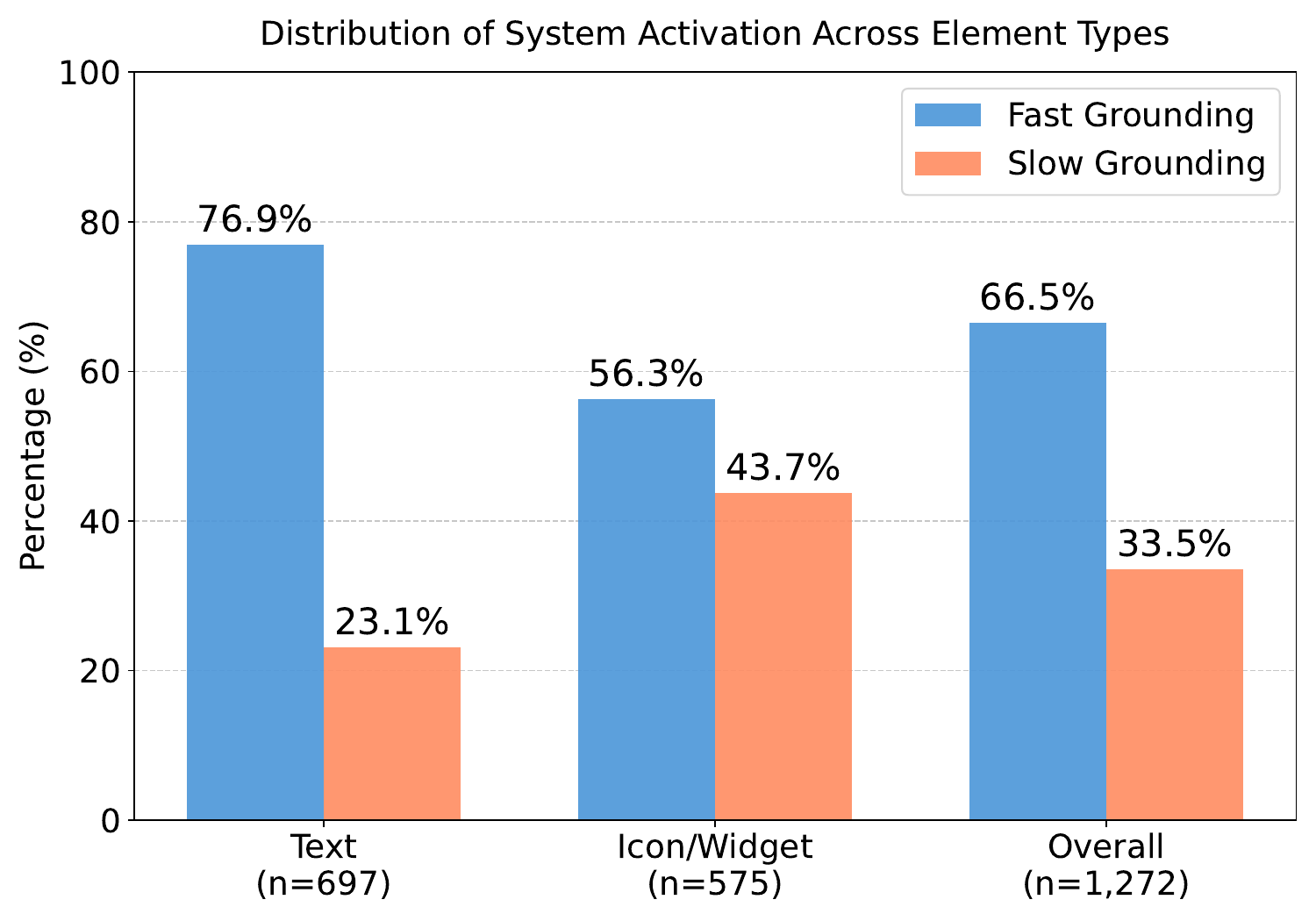} 
    \caption{Distribution of fast and slow system activation across different element types in \textbf{ScreenSpot}.} 
    \label{fig:thinking_count} 
\end{figure}

\textsc{Focus} achieved strong performance on the ScreenSpot benchmark through its dynamic system switching mechanism. However, we discovered that when the system exclusively relies on the slow thinking system ($\alpha$=1.0), as shown in Table \ref{fig:alpha_exp}, performance dramatically drops to 73.4\% while processing time increases from 2.6s to 5.4s. This performance degradation may result from \textbf{overthinking}: in simple scenarios where quick pattern recognition would suffice, the additional analytical processing steps can introduce noise into the decision process and lead to suboptimal predictions. The phenomenon also noted in recent research \cite{chen2025think23overthinkingo1like}. Moreover, the increased computational overhead significantly impacts system efficiency, making it impractical for real-world applications. The optimal configuration ($\alpha$=0.6) achieves the best trade-off, allowing \textsc{Focus} to adapt its processing strategy based on task complexity while maintaining reasonable computational costs. To further analyze system activation patterns, we examined the frequency of fast and slow grounding on ScreenSpot, as shown in Figure \ref{fig:thinking_count}. Text elements predominantly triggered fast grounding (76.9\%), benefiting from clear textual cues. In contrast, icons and widgets activated slow grounding more often (43.7\%), likely due to the absence of explicit text, requiring additional reasoning. Overall, the fast system was used 66.5\% of the time, demonstrating the effectiveness of adaptive switching in balancing efficiency and accuracy.
\subsection{Case Study}
To qualitatively demonstrate \textsc{{Focus}}'s advantages in handling complex UI scenarios, we conducted case study comparisons with ShowUI, as illustrated in Figure \ref{fig:case_study}. For additional cases and detailed reasoning analysis, please refer to Section \ref{sec:case_study}.
Consider the example in Figure \ref{fig:case_study}, where the task is to locate the \textbf{"View my account"} button on a GitLab interface. This case presents a complex layout with multiple similar interactive elements and hierarchical menu structures. ShowUI's direct grounding approach fails by incorrectly identifying project menu items as user profile controls. In contrast, \textsc{Focus} successfully activates its slow system upon detecting similar UI elements in the navigation bar (multiple numbered badges), enabling systematic analysis for precise localization, as shown in Table \ref{tab:focus_analysis}. Through its dual-system architecture, \textsc{Focus} achieves reliable grounding by leveraging comprehensive interface understanding.
\section{Related Work}

Recent years have witnessed significant advancement in GUI automation driven by large language models (LLMs). Early GUI agents predominantly focused on web interactions \cite{nakano2022webgptbrowserassistedquestionansweringhuman, hong2024cogagentvisuallanguagemodel} and have gradually expanded to mobile \cite{zhang2023appagentmultimodalagentssmartphone, wang2024mobileagentautonomousmultimodalmobile} and desktop environments \cite{zhang2024ufouifocusedagentwindows}. A fundamental challenge across these applications is precise element localization. Traditional approaches relied on structured information like XML and DOM trees \cite{zhang2023appagentmultimodalagentssmartphone}, but faced limitations in accessibility and information redundancy. Alternative methods using OCR \cite{du2020ppocrpracticalultralightweight} or detection models \cite{liu2024groundingdinomarryingdino} introduced additional computational overhead. Recent advances in multimodal large language models (MLLMs) have enabled direct GUI element localization \cite{hong2024cogagentvisuallanguagemodel, cheng2024seeclickharnessingguigrounding, lin2024showuivisionlanguageactionmodelgui}, partially bridging the visual perception gap. However, these approaches typically employ direct prediction without systematic analysis of interface structures. Our work introduces a dual-system framework combining fast prediction with systematic analysis for robust GUI grounding. See a comprehensive review of related work in Appendix \ref{sec:related_work}.
\section{Conclusion}
In this paper, we introduced \textsc{Focus}, a GUI grounding model that draws inspiration from human dual-process theory to combine fast prediction with deeper analysis. \textsc{Focus} incorporates two key innovations: (1) a dual-system architecture that simulates human cognitive mechanisms by decomposing grounding into progressive stages: from rapid intuitive judgments to in-depth analytical processing, and (2) an adaptive switching mechanism that balances efficiency and accuracy by transitioning between fast and slow systems based on task complexity. Extensive experiments on ScreenSpot and ScreenSpot-Pro benchmarks demonstrate that \textsc{Focus} achieves best performance across model scales (2B and larger), achieving 77.4\% and 13.3\% average accuracy respectively on these benchmarks.
\section*{Limitations}
The \textsc{{Focus}} framework faces several key limitations. The model struggles with icon recognition compared to text elements, particularly in professional software interfaces. The dual-system approach introduces additional computational overhead that may impact real-time performance. Moreover, the current model requires more sophisticated visual reasoning capabilities to better handle complex GUI scenarios, especially in specialized software environments.
\bibliography{custom}
\clearpage
\appendix
\section{Appendix}
\subsection{Related Work}
\label{sec:related_work}
\textbf{GUI Agents}
In recent years, large language models (LLMs) have achieved breakthrough advancements in language understanding and cognitive capabilities, driving the emergence of general-purpose intelligent agents like HuggingGPT \cite{shen2023hugginggptsolvingaitasks} and MetaGPT \cite{hong2024metagptmetaprogrammingmultiagent}. Extending this agent paradigm to real-world interfaces, researchers have begun applying LLMs to graphical user interface (GUI) automation across diverse domains. In web interaction, models like WebGPT \cite{nakano2022webgptbrowserassistedquestionansweringhuman}, CogAgent \cite{hong2024cogagentvisuallanguagemodel}, and AutoWebGLM \cite{lai2024autowebglmlargelanguagemodelbased} have demonstrated exceptional performance in navigation tasks. For mobile devices, the AppAgent series \cite{zhang2023appagentmultimodalagentssmartphone, li2024appagentv2advancedagent}, Mobile-Agent series \cite{wang2024mobileagentautonomousmultimodalmobile, wang2024mobileagentv2mobiledeviceoperation, wang2025mobileagenteselfevolvingmobileassistant}, and ScreenAgent \cite{niu2024screenagentvisionlanguagemodeldriven} have achieved effective interaction without requiring API access. In desktop environments, systems like UFO \cite{zhang2024ufouifocusedagentwindows} and OS-Copilot \cite{wu2024oscopilotgeneralistcomputeragents} successfully handle Windows tasks through coordinated multi-agent architectures. A critical challenge underlying all these GUI applications is the need to  comprehensively understand screen content and precisely locate interactive elements. Based on this foundational requirement, our work focuses on advancing element localization capabilities for GUI agents. \\
\textbf{Large Multimodal Models for Grounding}
Unlike traditional agents such as HuggingGPT \cite{shen2023hugginggptsolvingaitasks} and MetaGPT \cite{hong2024metagptmetaprogrammingmultiagent}, GUI Agents face a unique challenge: the need to accurately locate and understand interactive elements within graphical interfaces. Early GUI Agents primarily relied on structural information from XML and DOM trees \cite{zhang2023appagentmultimodalagentssmartphone, li2024appagentv2advancedagent}, which often suffer from information redundancy and, in many scenarios, these structured data may be unavailable or difficult to obtain. Alternative approaches utilizing tools like OCR \cite{du2020ppocrpracticalultralightweight} and SoM \cite{yang2023setofmarkpromptingunleashesextraordinary}, or detection models such as Grounding DINO \cite{liu2024groundingdinomarryingdino}, while effective, introduce non-end-to-end pipelines and additional computational overhead. With the rapid advancement of MLLMs \cite{you2024ferretuigroundedmobileui,wu2024deepseekvl2mixtureofexpertsvisionlanguagemodels}, researchers have begun exploring direct GUI element localization using MLLMs. For instance, models like CogAgent \cite{hong2024cogagentvisuallanguagemodel}, SeeClick \cite{cheng2024seeclickharnessingguigrounding}, ShowUI \cite{lin2024showuivisionlanguageactionmodelgui}, UGround \cite{gou2024navigatingdigitalworldhumans}, CoCo-Agent \cite{ma2024cocoagentcomprehensivecognitivemllm}, and InfiGUIAgent \cite{liu2025infiguiagentmultimodalgeneralistgui} have been trained on large-scale GUI datasets to directly localize elements from visual inputs, partially bridging the visual perception gap. Our work builds upon these MLLM-based approaches, but introduces a novel dual-system cognitive mechanism to enhance GUI grounding precision.
\subsection{Ablation Studies On $\alpha$}
\label{sec:alpha_exp}
The scaling factor $\alpha$ determines the dynamic balance between \textsc{Focus}'s fast and slow cognitive systems, with $\alpha = 0$ representing pure fast grounding and $\alpha = 1.0$ indicating exclusive use of slow cognitive processing. As shown in Figure \ref{fig:alpha_exp},Without the adaptive switching mechanism (w/o $\alpha$), the model automatically switches between systems based on initial token probabilities, achieving 74.8\% accuracy but with longer average processing times (3.2s). When  $\alpha$ = 0, the system depends entirely on the fast grounding system. Although this system is efficient, it has difficulty dealing with complex scenarios that require a deeper understanding of the scene and more detailed visual focused analysis.

As $\alpha$ increases to 0.6, we observe consistent performance improvements across all platforms, achieving optimal average accuracy of 77.4\% with a reasonable processing time of 2.6s, representing a +2.6\% improvement over the non-adaptive baseline. However, setting $\alpha = 1.0$ to exclusively use the slow cognitive system proves counterproductive, leading to decreased performance (73.4\% average accuracy) and significantly increased computational overhead (5.4s). This performance degradation occurs because excessive reasoning processes may introduce unnecessary complexity and longer processing times, even for simple tasks that could be efficiently handled by the fast grounding system. These results demonstrate that the optimal strategy lies in maintaining a balanced dual-system approach (0.4 $\leq \alpha \leq 0.6$), allowing \textsc{Focus} to adaptively switch between fast and slow processing based on task complexity.
\begin{figure*}[t!]
    \centering
    \begin{subfigure}[t]{0.48\textwidth}
        \centering
        \includegraphics[width=\textwidth]{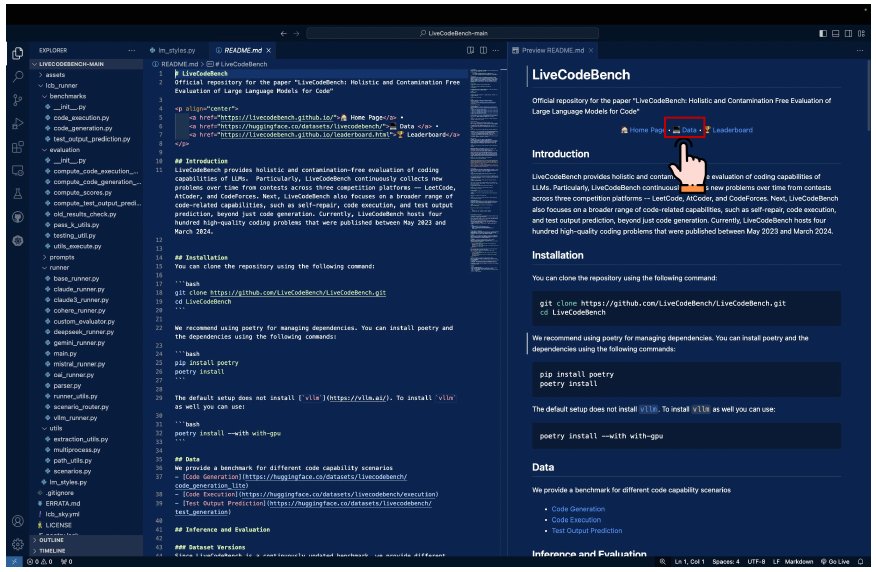}
        \caption{\ding{52} \textsc{\textbf{Focus}} made a correct prediction in locating the data download link.}
        \label{fig:app_focus_case}
    \end{subfigure}
    \hfill
    \begin{subfigure}[t]{0.48\textwidth}
        \centering
        \includegraphics[width=\textwidth]{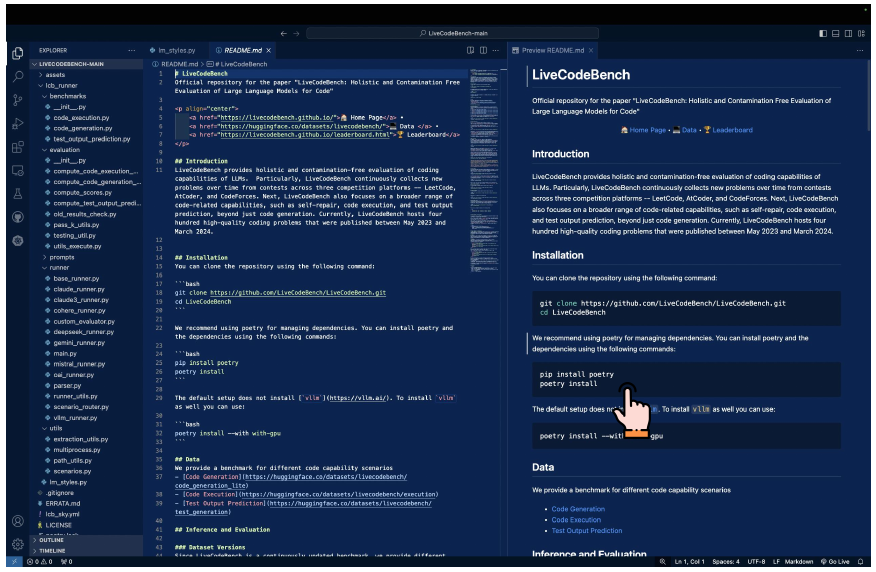}
        \caption{\ding{55} \textbf{ShowUI} \cite{lin2024showuivisionlanguageactionmodelgui} failed to locate the data download link.}
        \label{fig:app_showui_case}
    \end{subfigure}
    \begin{subfigure}[t]{0.48\textwidth}
        \centering
        \includegraphics[width=\textwidth]{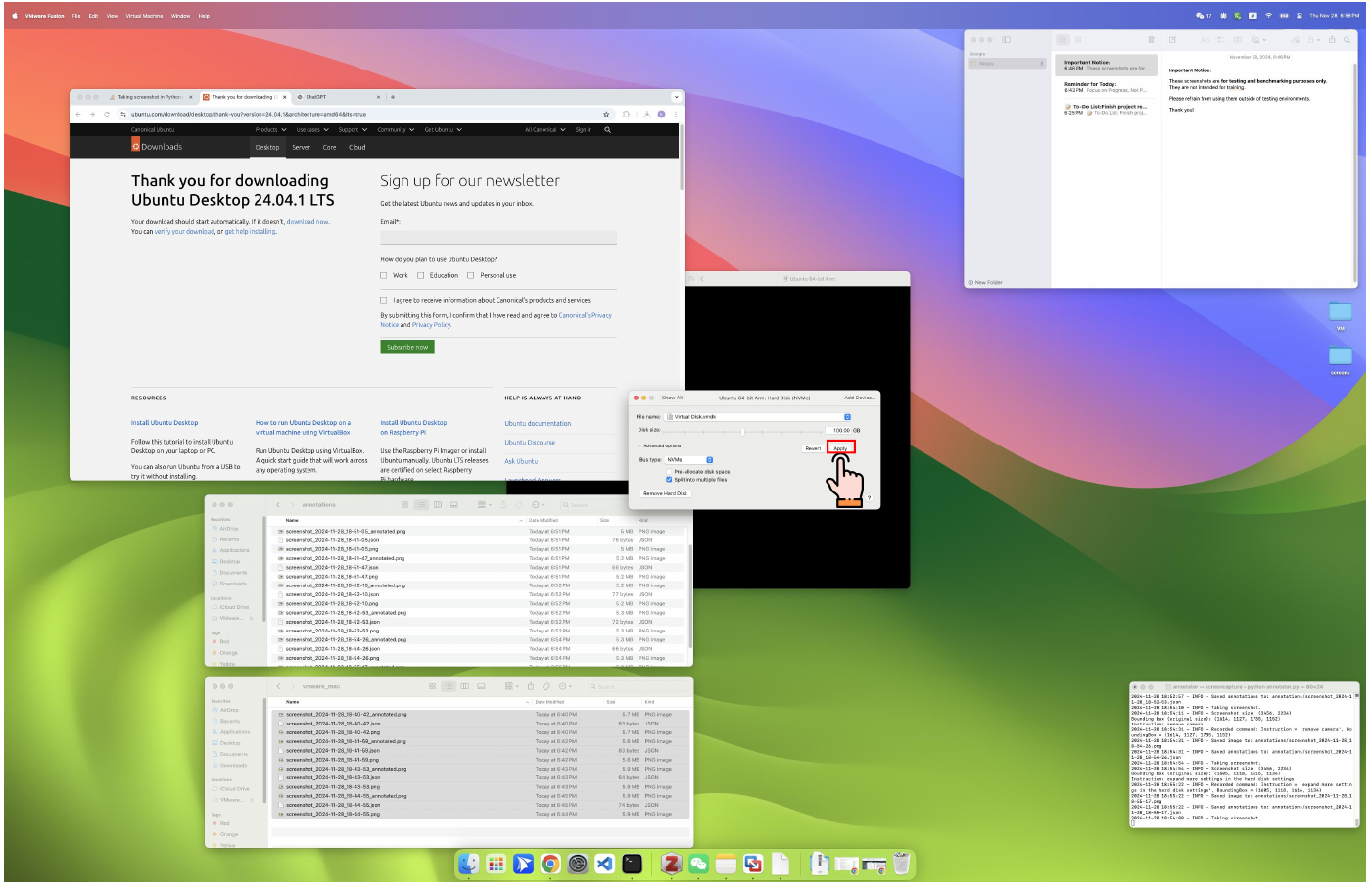}
        \caption{\ding{52} \textsc{\textbf{Focus}} successfully identified the disk settings button.}
        \label{fig:app_focus_case_2}
    \end{subfigure}
    \hfill
    \begin{subfigure}[t]{0.48\textwidth}
        \centering
        \includegraphics[width=\textwidth]{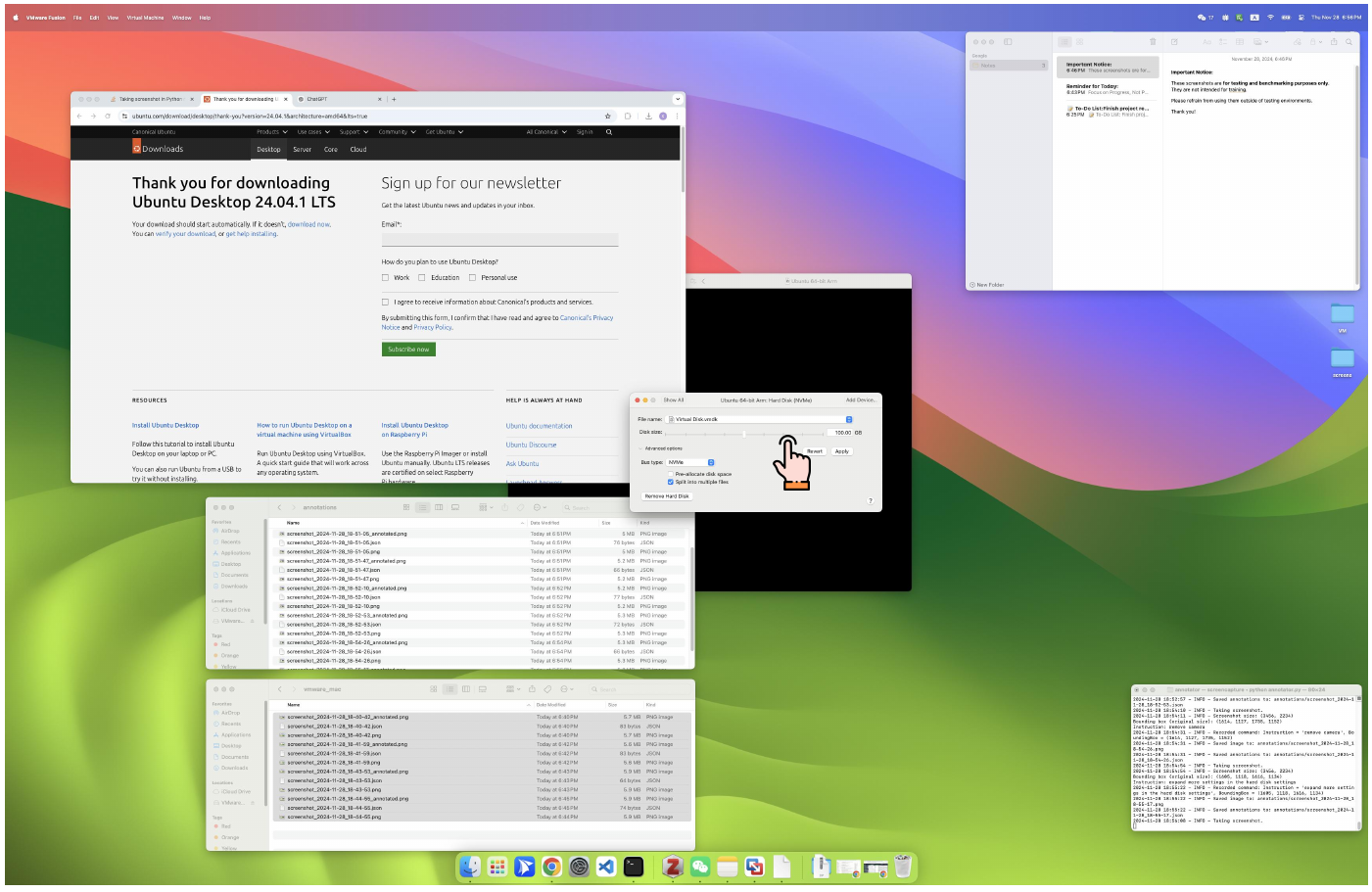}
        \caption{\ding{55} \textbf{ShowUI} \cite{lin2024showuivisionlanguageactionmodelgui} failed to locate the disk settings button.}
        \label{fig:app_showui_case_2}
    \end{subfigure}
    \caption{Case study comparison between ShowUI and \textsc{Focus}. For cases (a) and (b), the instruction is \textbf{"Download the LiveCodeBench's data"}, where \textsc{{Focus}} successfully locates the target through comprehensive interface analysis while ShowUI struggles with similar elements. For cases (c) and (d), the instruction is \textbf{"apply the new disk settings"}, demonstrating \textsc{{Focus}}'s consistent performance in complex layouts compared to ShowUI's limitations in handling diverse interface scenarios.}
    \label{fig:app_case_study}
\end{figure*}
\subsection{Details On Evaluation GUI Grounding Benchmark}
\label{sec:inference-details}
\begin{figure}[htbp] 
    \centering
\includegraphics[width=0.48\textwidth]{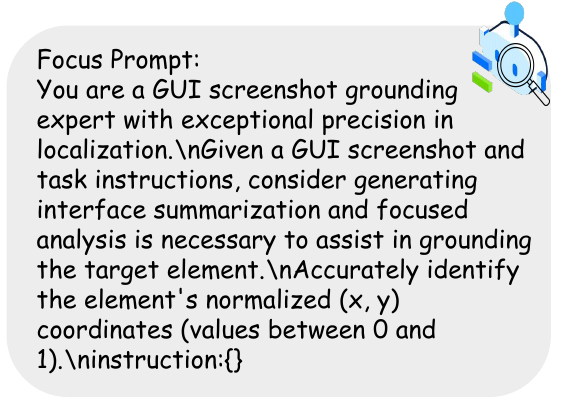} 
    \caption{\textsc{{Focus}} uses prompt template for GUI grounding. The prompt instructs the model to act as a GUI screenshot grounding expert and generate normalized coordinates for target elements.} 
    \label{fig:prompt}  
\end{figure}

\textbf{Evaluation Details.} When evaluating Screenspot and Screenspot-pro, we used prompts as illustrated in Figure \ref{fig:prompt}. The model deployment was based on the Transformers \cite{wolf2020huggingfacestransformersstateoftheartnatural} framework, with image resolutions ranging from 256x28x28 to 1344x28x28. We loaded the qwen2-vl model in bf16 format, with a maximum output limit of 4096 tokens. 
\\
During inference, we set \( \alpha \) to 0.6 to achieve optimal balance between fast and slow systems. All evaluation experiments used a single A100 (40GB) GPU.
\textbf{Compared Method.} Recently, Many GUI grounding method based on MLLMs has been proposed, and we have selected the following methods as comparisons for \textsc{Focus}:
\begin{itemize}
\item \textbf{Fuyu} supports arbitrary image resolutions and excels at graph, diagram, and UI-based tasks through its streamlined multimodal architecture \cite{bavishi2023introducing}.
\item \textbf{CogAgent} integrates visual-language understanding with a specialized GUI action space to enable efficient interface interactions \cite{hong2024cogagentvisuallanguagemodel}.
\item \textbf{SeeClick} employs Qwen-VL-Chat \cite{bai2023qwenvlversatilevisionlanguagemodel} as its visual backbone and leverages hierarchical visual grounding for advanced GUI interactions, focusing on precise element localization \cite{cheng2024seeclickharnessingguigrounding}.
\item \textbf{UGround} aims to replicate human-like navigation patterns in GUI through universal visual grounding strategies \cite{gou2024navigatingdigitalworldhumans}.
\item \textbf{ShowUI} builds upon Qwen2-VL-2B \cite{wang2024qwen2vlenhancingvisionlanguagemodels} with UI-guided visual token selection and interleaved vision-language-action streaming for comprehensive GUI task handling \cite{lin2024showuivisionlanguageactionmodelgui}.
\item \textbf{InfiGUIAgent} utilizes Qwen2-VL-2B \cite{wang2024qwen2vlenhancingvisionlanguagemodels} as its backbone and incorporates native reasoning and reflection mechanisms for GUI interaction \cite{liu2025infiguiagentmultimodalgeneralistgui}.
\item \textbf{Aria-UI} leverages the SoTA multimodal MoE model Aria and specializes in visual grounding of GUI instructions \cite{yang2024ariauivisualgroundinggui}.
\item \textbf{OS-Atlas} employs InternVL-2-4B \cite{chen2024internvlscalingvisionfoundation} as its foundation and provides a foundation action model for cross-platform GUI interactions \cite{wu2024osatlasfoundationactionmodel}.
\end{itemize}

We selected a comprehensive range of baseline methods that utilize comparable or larger model sizes and similar backbone architectures, including methods built upon the same Qwen2-VL-2B foundation as \textsc{Focus}. This diverse selection of state-of-the-art methods, spanning from specialized GUI agents to general-purpose multimodal models, ensures a thorough and fair evaluation of our approach.
\subsection{Case Studies On \textsc{{Focus}}}
We conducted case studies to demonstrate the effectiveness of Focus compared to fast grounding methods. As shown in Figure \ref{fig:app_case_study}(a-d), when dealing with the LiveCodeBench repository interface, ShowUI fails to accurately locate the data download element due to several challenges: (1) the presence of multiple similar clickable elements in the interface, (2) the hierarchical structure of the documentation page, and (3) the need to understand the semantic relationship between the instruction and interface elements.
\begin{table}[htp]
\small
\centering
\begin{tabular}{p{0.95\linewidth}}
\toprule
\textbf{Analysis Process for LiveCodeBench Data Download} \\
\midrule
\textit{<|summary\_start|>} The screen displays a repository page for LiveCodeBench, with a focus on the 'Data' section. The 'Data' section provides links to download the repository's data and documentation. \textit{<|summary\_end|>} \\
\textit{<|Focus\_start|>} The 'Data' section is located in the upper right corner of the screen, under the 'Introduction' heading. It contains two links: 'Home Page' and 'Data', which are clickable and lead to the repository's documentation and data download page. \textit{<|Focus\_end|>} \\
\bottomrule
\end{tabular}
\caption{\textsc{{Focus}}'s systematic analysis process for the LiveCodeBench interface.}
\label{tab:focus_analysis}
\end{table}
\begin{table}[htp]
\small
\centering
\begin{tabular}{p{0.95\linewidth}}
\toprule
\textbf{Analysis Process for GitLab Project Interface} \\
\midrule
\textit{<|summary\_start|>} The screen displays a GitLab projects dashboard interface, showing a list of repositories with their associated metadata. The interface includes a navigation bar at the top with search functionality and user controls, and the main content area lists various project repositories with their details and metrics. \textit{<|summary\_end|>} \\
\textit{<|Focus\_start|>} The user account access control is positioned in the top-right corner of the navigation bar. It appears as a circular profile avatar with the number "13", indicating notifications or updates. This element is separate from other navigation items like search, project metrics (showing "21"), and merge requests (showing "20"), making it a distinct interactive element for accessing account-related functions. \textit{<|Focus\_end|>} \\
\bottomrule
\end{tabular}
\caption{\textsc{{Focus}}'s systematic analysis process for the GitLab interface.}
\label{tab:focus_analysis}
\end{table}
In contrast, \textsc{{Focus}} adopts a systematic analysis approach, as detailed in Table \ref{tab:focus_analysis}. Through its dual-system mechanism, \textsc{{Focus}} first comprehends the overall interface layout, recognizing it as a LiveCodeBench repository page with a dedicated 'Data' section for downloading repository data and documentation. It then conducts detailed visual analysis, identifying that the target `Data' section is specifically positioned in the upper right corner under the `Introduction' heading, containing clickable links for documentation and data download.

This case demonstrates how \textsc{Focus}'s slow grounding system enables accurate element localization through comprehensive interface understanding and detailed visual focused analysis, overcoming the limitations of pure fast grounding methods that struggle with complex layouts and similar elements. Through the dual-system architecture, \textsc{Focus} first generates a high-level interface summarization to establish global context, followed by targeted visual analysis of specific regions of interest. This hierarchical processing approach proves particularly effective when handling ambiguous scenarios, such as interfaces with multiple similar interactive elements or nested menu structures. The success in these challenging cases validates our hypothesis that combining fast intuitive processing with deliberate analytical reasoning better mirrors human cognitive patterns in GUI interaction. Furthermore, the consistent performance across diverse interface scenarios suggests that this dual-system approach could be extended to other visual grounding tasks requiring both efficiency and precision.
\label{sec:case_study}
\label{sec:appendix}
\end{document}

%% file: tables/dataset_count.tex
\begin{table}[]
\centering
\begin{adjustbox}{width=0.48\textwidth} 
\begin{tabular}{cllll}
\hline
\multicolumn{1}{l}{Usage}                                                    & Source                                                                                                                  & Number & \textbf{\#S\_Num} & \textbf{\#F\_Num} \\ \hline
\multirow{4}{*}{\begin{tabular}[c]{@{}c@{}} \\ \\ Single \\ Grounding\end{tabular}} & \begin{tabular}[c]{@{}l@{}}Wave-UI \\ \cite{zheng2024agentstudio}\end{tabular}                         & 36K    & 15K               & 21K               \\
                                                                             & \begin{tabular}[c]{@{}l@{}}GUICourse \\ \cite{chen2024guicoursegeneralvisionlanguage}\end{tabular}     & 170K   & 100K              & 70K               \\
                                                                             & \begin{tabular}[c]{@{}l@{}}Aguvis-stage1 \\ \cite{xu2024aguvis}\end{tabular}                           & 80K    & 27K               & 53K               \\
                                                                             & \begin{tabular}[c]{@{}l@{}}Desktop-UI \\ \cite{lin2024showuivisionlanguageactionmodelgui}\end{tabular} & 8K     & 3K                & 5K                \\ \hline
\multicolumn{1}{l}{Total}                                                    &                                                                                                                         & 300K   & 145K              & 150K              \\ \hline
\end{tabular}
\end{adjustbox}
\caption{Dataset Statistics. \textbf{Single Grounding} refers to single-step interaction samples without previous steps. \textbf{\#S\_Num} and \textbf{\#F\_Num} represent the number of samples in Slow Grounding and Fast Grounding data.}
\label{tab:dataset_count}
\end{table}

%% file: tables/screenspot.tex
\begin{table*}[t!]
\centering
\small
\renewcommand\arraystretch{1.1}
\tabcolsep=0.05cm
{\fontsize{10pt}{12pt}\selectfont
\begin{tabular}{ccccccccc}
\hline
\multirow{2}{*}{\makecell{GUI Agent \\ MLLMs}} & \multirow{2}{*}{\parbox{1.5cm}{\centering Model\\Size}} & \multicolumn{2}{c}{Mobile} & \multicolumn{2}{c}{Desktop} & \multicolumn{2}{c}{Web} & \multirow{2}{*}{Average} \\ \cline{3-8}
 & & Text & Icon/Widget & Text & Icon/Widget & Text & Icon/Widget & \\
\hline
GPT-4          & -    & 22.6 & 24.5 & 20.2 & 11.8 & 9.2 & 8.8 & 16.7 \\
GPT-4o         & -    & 20.2 & 24.9 & 21.1 & 23.6 & 12.2 & 7.8 & 18.1 \\
Gemini-1.5-pro & -    & 76.2 & 54.1 & 65.5 & 39.2 & 52.2 & 32.0 & 53.2 \\
\hline
Qwen2-VL-2B    & 2B   & 24.2 & 10.0 & 1.4 & 9.3 & 8.7 & 2.41 & 9.3 \\
Qwen2-VL-7B    & 7B   & 61.3 & 39.3 & 52.0 & 45.0 & 33.0 & 21.8 & 42.9 \\
Fuyu           & 8B   & 41.0 & 1.3 & 33.0 & 3.6 & 33.9 & 4.4 & 19.5 \\
CogAgent       & 18B  & 67.0 & 24.0 & 74.2 & 20.0 & 70.4 & 28.6 & 47.4 \\
Seeclick       & 9.6B & 78.0 & 52.0 & 72.2 & 30.0 & 55.7 & 32.5 & 53.4 \\
UGround        & 7B   & 82.8 & 60.3 & 82.5 & 63.6 & 80.4 & 70.4 & 73.3 \\
ShowUI         & 2B   & 92.3 & 75.5 & 76.3 & 61.1 & 81.7 & 63.6 & 75.1 \\
InfiGUIAgent   & 2B   & 88.6 & 74.7 & 85.6 & 65.0 & 79.1 & 64.6 & 76.3 \\
\hline
\textbf{\textsc{Focus} } (ours)      & 2B   & 90.1 & 78.2 & 80.9 & 65.0 & 81.7 & 68.5 & 
\textbf{77.4} \\
\hline
\end{tabular}
}
\caption{GUI Grounding Results of different GUI Agent on \textbf{ScreenSpot}.}
\label{tab:screenspot}
\vspace{-1.0em}
\end{table*}

%% file: tables/screenspot-pro.tex
\begin{table*}[htbp]
\centering
\small
\centering
\renewcommand\arraystretch{1.1}
\tabcolsep=0.05cm
{\fontsize{10pt}{12pt}\selectfont
\resizebox{\textwidth}{!}{
\begin{tabular}{l|ccc|ccc|ccc|ccc|ccc|ccc|ccc}
\toprule
\multirow{2}{*}{\textbf{Model}} & \multicolumn{3}{c|}{\textbf{Development}} & \multicolumn{3}{c|}{\textbf{Creative}} & \multicolumn{3}{c|}{\textbf{CAD}} & \multicolumn{3}{c|}{\textbf{Scientific}} & \multicolumn{3}{c|}{\textbf{Office}} & \multicolumn{3}{c|}{\textbf{OS}} & \multicolumn{3}{c}{\textbf{Overall}} \\
& \textbf{Text} & \textbf{Icon} & \textbf{Avg} & \textbf{Text} & \textbf{Icon} & \textbf{Avg} & \textbf{Text} & \textbf{Icon} & \textbf{Avg} & \textbf{Text} & \textbf{Icon} & \textbf{Avg} & \textbf{Text} & \textbf{Icon} & \textbf{Avg} & \textbf{Text} & \textbf{Icon} & \textbf{Avg} & \textbf{Text} & \textbf{Icon} & \textbf{Avg} \\

\midrule
AriaUI (MOE, 3.9B active) & 16.2 & 0.0 & 8.4 & 23.7 & 2.1 & 14.7 & 7.6 & 1.6 & 6.1 & 27.1 & 6.4 & 18.1 & 20.3 & 1.9 & 16.1 & 4.7 & 0.0 & 2.6 & 17.1 & 2.0 & 11.3 \\
CogAgent (18B) & 14.9 & 0.7 & 8.0 & 9.6 & 0.0 & 5.6 & 7.1 & 3.1 & 6.1 & 22.2 & 1.8 & 13.4 & 13.0 & 0.0 & 10.0 & 5.6 & 0.0 & 3.1 & 12.0 & 0.8 & 7.7 \\
ShowUI (2B) & 16.9 & 1.4 & 9.4 & 9.1 & 0.0 & 5.3 & 2.5 & 0.0 & 1.9 & 13.2 & 7.3 & 10.6 & 15.3 & 7.5 & 13.5 & 10.3 & 2.2 & 6.6 & 10.8 & 2.6 & 7.7 \\
OSAtlas-4B & 7.1 & 0.0 & 3.7 & 3.0 & 1.4 & 2.3 & 2.0 & 0.0 & 1.5 & 9.0 & 5.5 & 7.5 & 5.1 & 3.8 & 4.8 & 5.6 & 0.0 & 3.1 & 5.0 & 1.7 & 3.7 \\
MiniCPM-V (7B) & 7.1 & 0.0 & 3.7 & 2.0 & 0.0 & 1.2 & 4.1 & 1.6 & 3.4 & 8.3 & 0.0 & 4.7 & 2.8 & 3.8 & 3.0 & 3.7 & 1.1 & 2.6 & 4.5 & 0.7 & 3.0 \\
Qwen2-VL-7B & 2.6 & 0.0 & 1.3 & 1.5 & 0.0 & 0.9 & 0.5 & 0.0 & 0.4 & 6.3 & 0.0 & 3.5 & 3.4 & 1.9 & 3.0 & 0.9 & 0.0 & 0.5 & 2.5 & 0.2 & 1.6 \\
SeeClick (7B) & 0.6 & 0.0 & 0.3 & 1.0 & 0.0 & 0.6 & 2.5 & 0.0 & 1.9 & 3.5 & 0.0 & 2.0 & 1.1 & 0.0 & 0.9 & 2.8 & 0.0 & 1.5 & 1.8 & 0.0 & 1.1 \\
GPT-4o & 1.3 & 0.0 & 0.7 & 1.0 & 0.0 & 0.6 & 2.0 & 0.0 & 1.5 & 2.1 & 0.0 & 1.2 & 1.1 & 0.0 & 0.9 & 0.0 & 0.0 & 0.0 & 1.3 & 0.0 & 0.8 \\
QwenVL-7B & 0.0 & 0.0 & 0.0 & 0.0 & 0.0 & 0.0 & 0.0 & 0.0 & 0.0 & 0.7 & 0.0 & 0.4 & 0.0 & 0.0 & 0.0 & 0.0 & 0.0 & 0.0 & 0.1 & 0.0 & 0.1 \\
\textbf{\textsc{Focus} (ours)} & 22.8 & 1.7 & 12.4 & 23.7 & 1.5 & 14.4 & 7.6 & 3.1 & 6.5 & 25.0 & 7.1 & 16.9 & 23.2 & 7.7 & 19.1 & 17.8 & 2.5 & 10.7 & \textbf{19.8} & \textbf{3.9} & \textbf{13.3} \\
\hline
\end{tabular}
}}
\caption{Performance breakdown of various models across application categories on ScreenSpot-Pro.}
\label{tab:result-by-group-and-ui-type}
\end{table*}

%% file: tables/ablation_study.tex
\begin{table*}[]
\centering
\renewcommand\arraystretch{1.1}
{\fontsize{9pt}{12pt}\selectfont
\begin{tabular}{lccccccccc}
\hline
\multirow{2}{*}{Model Variant} & \multirow{2}{*}{Summary} & \multirow{2}{*}{Focus} & \multicolumn{2}{c}{Mobile} & \multicolumn{2}{c}{Desktop} & \multicolumn{2}{c}{Web} & \multirow{2}{*}{Average} \\ \cline{4-9}
                                        &                          &                        & Text     & Icon/Widget     & Text      & Icon/Widget     & Text    & Icon/Widget   &                          \\ \hline
\textsc{Focus}                          & \ding{52}              & \ding{52}           & 90.1     & 78.2            & 80.9      & 65.0            & 81.7    & 68.5          & \textbf{77.4}            \\
w/o Both                                & \ding{55}                 & \ding{55}               & 84.2     & 72.5            & 75.3      & 60.1            & 74.8    & 61.2          & 71.4                     \\
w/o Summary                             & \ding{55}                & \ding{52}           & 87.3     & 75.8            & 77.8      & 62.5            & 78.2    & 63.6          & 74.2                     \\
w/o Focus                               & \ding{52}             & \ding{55}              & 86.5     & 74.9            & 76.9      & 61.8            & 77.5    & 62.9          & 73.4                     \\ \hline
\end{tabular}
}
\caption{Ablation study of \textsc{Focus} on ScreenSpot dataset. \ding{52}  indicates the component is included, while \ding{55} indicates it is removed. All numbers are reported as percentages. Results demonstrate that both interface summarization and focused analysis components contribute substantially to model performance.}
\label{tab:ablation}
\end{table*}

%% file: acl_latex.bbl
\begin{thebibliography}{43}
\providecommand{\natexlab}[1]{#1}

\bibitem[{Bai et~al.(2023)Bai, Bai, Yang, Wang, Tan, Wang, Lin, Zhou, and Zhou}]{bai2023qwenvlversatilevisionlanguagemodel}
Jinze Bai, Shuai Bai, Shusheng Yang, Shijie Wang, Sinan Tan, Peng Wang, Junyang Lin, Chang Zhou, and Jingren Zhou. 2023.
\newblock \href {https://arxiv.org/abs/2308.12966} {Qwen-vl: A versatile vision-language model for understanding, localization, text reading, and beyond}.
\newblock \emph{Preprint}, arXiv:2308.12966.

\bibitem[{Bavishi et~al.(2023)Bavishi, Elsen, Hawthorne, Nye, Odena, Somani, and Ta\c{s}{\i}rlar}]{bavishi2023introducing}
Rohan Bavishi, Erich Elsen, Curtis Hawthorne, Maxwell Nye, Augustus Odena, Arushi Somani, and Sagnak Ta\c{s}{\i}rlar. 2023.
\newblock Introducing our multimodal models.

\bibitem[{Chen et~al.(2024{\natexlab{a}})Chen, Cui, Hu, Qin, Fang, Zhao, Wang, Liu, Chen, Huo, Yao, Lin, Liu, and Sun}]{chen2024guicoursegeneralvisionlanguage}
Wentong Chen, Junbo Cui, Jinyi Hu, Yujia Qin, Junjie Fang, Yue Zhao, Chongyi Wang, Jun Liu, Guirong Chen, Yupeng Huo, Yuan Yao, Yankai Lin, Zhiyuan Liu, and Maosong Sun. 2024{\natexlab{a}}.
\newblock \href {https://arxiv.org/abs/2406.11317} {Guicourse: From general vision language models to versatile gui agents}.
\newblock \emph{Preprint}, arXiv:2406.11317.

\bibitem[{Chen et~al.(2025)Chen, Xu, Liang, He, Pang, Yu, Song, Liu, Zhou, Zhang, Wang, Tu, Mi, and Yu}]{chen2025think23overthinkingo1like}
Xingyu Chen, Jiahao Xu, Tian Liang, Zhiwei He, Jianhui Pang, Dian Yu, Linfeng Song, Qiuzhi Liu, Mengfei Zhou, Zhuosheng Zhang, Rui Wang, Zhaopeng Tu, Haitao Mi, and Dong Yu. 2025.
\newblock \href {https://arxiv.org/abs/2412.21187} {Do not think that much for 2+3=? on the overthinking of o1-like llms}.
\newblock \emph{Preprint}, arXiv:2412.21187.

\bibitem[{Chen et~al.(2024{\natexlab{b}})Chen, Wu, Wang, Su, Chen, Xing, Zhong, Zhang, Zhu, Lu, Li, Luo, Lu, Qiao, and Dai}]{chen2024internvlscalingvisionfoundation}
Zhe Chen, Jiannan Wu, Wenhai Wang, Weijie Su, Guo Chen, Sen Xing, Muyan Zhong, Qinglong Zhang, Xizhou Zhu, Lewei Lu, Bin Li, Ping Luo, Tong Lu, Yu~Qiao, and Jifeng Dai. 2024{\natexlab{b}}.
\newblock \href {https://arxiv.org/abs/2312.14238} {Internvl: Scaling up vision foundation models and aligning for generic visual-linguistic tasks}.
\newblock \emph{Preprint}, arXiv:2312.14238.

\bibitem[{Cheng et~al.(2024)Cheng, Sun, Chu, Xu, Li, Zhang, and Wu}]{cheng2024seeclickharnessingguigrounding}
Kanzhi Cheng, Qiushi Sun, Yougang Chu, Fangzhi Xu, Yantao Li, Jianbing Zhang, and Zhiyong Wu. 2024.
\newblock \href {https://arxiv.org/abs/2401.10935} {Seeclick: Harnessing gui grounding for advanced visual gui agents}.
\newblock \emph{Preprint}, arXiv:2401.10935.

\bibitem[{Du et~al.(2020)Du, Li, Guo, Yin, Liu, Zhou, Bai, Yu, Yang, Dang, and Wang}]{du2020ppocrpracticalultralightweight}
Yuning Du, Chenxia Li, Ruoyu Guo, Xiaoting Yin, Weiwei Liu, Jun Zhou, Yifan Bai, Zilin Yu, Yehua Yang, Qingqing Dang, and Haoshuang Wang. 2020.
\newblock \href {https://arxiv.org/abs/2009.09941} {Pp-ocr: A practical ultra lightweight ocr system}.
\newblock \emph{Preprint}, arXiv:2009.09941.

\bibitem[{Evans(2008)}]{evans2008dual}
Jonathan St~BT Evans. 2008.
\newblock Dual-processing accounts of reasoning, judgment, and social cognition.
\newblock \emph{Annu. Rev. Psychol.}, 59(1):255--278.

\bibitem[{Gou et~al.(2024)Gou, Wang, Zheng, Xie, Chang, Shu, Sun, and Su}]{gou2024navigatingdigitalworldhumans}
Boyu Gou, Ruohan Wang, Boyuan Zheng, Yanan Xie, Cheng Chang, Yiheng Shu, Huan Sun, and Yu~Su. 2024.
\newblock \href {https://arxiv.org/abs/2410.05243} {Navigating the digital world as humans do: Universal visual grounding for gui agents}.
\newblock \emph{Preprint}, arXiv:2410.05243.

\bibitem[{He et~al.(2024)He, Jin, Xia, Su, Fan, Zou, Hu, and Liu}]{he2024pcagentsleepai}
Yanheng He, Jiahe Jin, Shijie Xia, Jiadi Su, Runze Fan, Haoyang Zou, Xiangkun Hu, and Pengfei Liu. 2024.
\newblock \href {https://arxiv.org/abs/2412.17589} {Pc agent: While you sleep, ai works -- a cognitive journey into digital world}.
\newblock \emph{Preprint}, arXiv:2412.17589.

\bibitem[{Hong et~al.(2024{\natexlab{a}})Hong, Zhuge, Chen, Zheng, Cheng, Zhang, Wang, Wang, Yau, Lin, Zhou, Ran, Xiao, Wu, and Schmidhuber}]{hong2024metagptmetaprogrammingmultiagent}
Sirui Hong, Mingchen Zhuge, Jiaqi Chen, Xiawu Zheng, Yuheng Cheng, Ceyao Zhang, Jinlin Wang, Zili Wang, Steven Ka~Shing Yau, Zijuan Lin, Liyang Zhou, Chenyu Ran, Lingfeng Xiao, Chenglin Wu, and Jürgen Schmidhuber. 2024{\natexlab{a}}.
\newblock \href {https://arxiv.org/abs/2308.00352} {Metagpt: Meta programming for a multi-agent collaborative framework}.
\newblock \emph{Preprint}, arXiv:2308.00352.

\bibitem[{Hong et~al.(2024{\natexlab{b}})Hong, Wang, Lv, Xu, Yu, Ji, Wang, Wang, Zhang, Li, Xu, Dong, Ding, and Tang}]{hong2024cogagentvisuallanguagemodel}
Wenyi Hong, Weihan Wang, Qingsong Lv, Jiazheng Xu, Wenmeng Yu, Junhui Ji, Yan Wang, Zihan Wang, Yuxuan Zhang, Juanzi Li, Bin Xu, Yuxiao Dong, Ming Ding, and Jie Tang. 2024{\natexlab{b}}.
\newblock \href {https://arxiv.org/abs/2312.08914} {Cogagent: A visual language model for gui agents}.
\newblock \emph{Preprint}, arXiv:2312.08914.

\bibitem[{Hu et~al.(2024)Hu, Ouyang, Gao, and Shou}]{hu2024dawnguiagentpreliminary}
Siyuan Hu, Mingyu Ouyang, Difei Gao, and Mike~Zheng Shou. 2024.
\newblock \href {https://arxiv.org/abs/2411.10323} {The dawn of gui agent: A preliminary case study with claude 3.5 computer use}.
\newblock \emph{Preprint}, arXiv:2411.10323.

\bibitem[{Kahneman(2011)}]{kahneman2011thinking}
Daniel Kahneman. 2011.
\newblock Thinking, fast and slow.
\newblock \emph{Farrar, Straus and Giroux}.

\bibitem[{Lai et~al.(2024)Lai, Liu, Iong, Yao, Chen, Shen, Yu, Zhang, Zhang, Dong, and Tang}]{lai2024autowebglmlargelanguagemodelbased}
Hanyu Lai, Xiao Liu, Iat~Long Iong, Shuntian Yao, Yuxuan Chen, Pengbo Shen, Hao Yu, Hanchen Zhang, Xiaohan Zhang, Yuxiao Dong, and Jie Tang. 2024.
\newblock \href {https://arxiv.org/abs/2404.03648} {Autowebglm: A large language model-based web navigating agent}.
\newblock \emph{Preprint}, arXiv:2404.03648.

\bibitem[{Li et~al.(2025)Li, Meng, Lin, Luo, Tian, Ma, Huang, and Chua}]{li2024screenspot-pro}
Kaixin Li, Ziyang Meng, Hongzhan Lin, Ziyang Luo, Yuchen Tian, Jing Ma, Zhiyong Huang, and Tat-Seng Chua. 2025.
\newblock Screenspot-pro: Gui grounding for professional high-resolution computer use.

\bibitem[{Li et~al.(2024)Li, Zhang, Yang, Fu, Cheng, Chen, Chen, and Wei}]{li2024appagentv2advancedagent}
Yanda Li, Chi Zhang, Wanqi Yang, Bin Fu, Pei Cheng, Xin Chen, Ling Chen, and Yunchao Wei. 2024.
\newblock \href {https://arxiv.org/abs/2408.11824} {Appagent v2: Advanced agent for flexible mobile interactions}.
\newblock \emph{Preprint}, arXiv:2408.11824.

\bibitem[{Lin et~al.(2024)Lin, Li, Gao, Yang, Wu, Bai, Lei, Wang, and Shou}]{lin2024showuivisionlanguageactionmodelgui}
Kevin~Qinghong Lin, Linjie Li, Difei Gao, Zhengyuan Yang, Shiwei Wu, Zechen Bai, Weixian Lei, Lijuan Wang, and Mike~Zheng Shou. 2024.
\newblock \href {https://arxiv.org/abs/2411.17465} {Showui: One vision-language-action model for gui visual agent}.
\newblock \emph{Preprint}, arXiv:2411.17465.

\bibitem[{Liu et~al.(2023)Liu, Li, Wu, and Lee}]{liu2023visualinstructiontuning}
Haotian Liu, Chunyuan Li, Qingyang Wu, and Yong~Jae Lee. 2023.
\newblock \href {https://arxiv.org/abs/2304.08485} {Visual instruction tuning}.
\newblock \emph{Preprint}, arXiv:2304.08485.

\bibitem[{Liu et~al.(2024)Liu, Zeng, Ren, Li, Zhang, Yang, Jiang, Li, Yang, Su, Zhu, and Zhang}]{liu2024groundingdinomarryingdino}
Shilong Liu, Zhaoyang Zeng, Tianhe Ren, Feng Li, Hao Zhang, Jie Yang, Qing Jiang, Chunyuan Li, Jianwei Yang, Hang Su, Jun Zhu, and Lei Zhang. 2024.
\newblock \href {https://arxiv.org/abs/2303.05499} {Grounding dino: Marrying dino with grounded pre-training for open-set object detection}.
\newblock \emph{Preprint}, arXiv:2303.05499.

\bibitem[{Liu et~al.(2025)Liu, Li, Wei, Xie, Hu, Xu, Zhang, Han, Yang, and Wu}]{liu2025infiguiagentmultimodalgeneralistgui}
Yuhang Liu, Pengxiang Li, Zishu Wei, Congkai Xie, Xueyu Hu, Xinchen Xu, Shengyu Zhang, Xiaotian Han, Hongxia Yang, and Fei Wu. 2025.
\newblock \href {https://arxiv.org/abs/2501.04575} {Infiguiagent: A multimodal generalist gui agent with native reasoning and reflection}.
\newblock \emph{Preprint}, arXiv:2501.04575.

\bibitem[{Lu et~al.(2024)Lu, Liu, Zhang, Wang, Dong, Liu, Sun, Ren, Li, Yang, Sun, Deng, Xu, Xie, and Ruan}]{lu2024deepseekvlrealworldvisionlanguageunderstanding}
Haoyu Lu, Wen Liu, Bo~Zhang, Bingxuan Wang, Kai Dong, Bo~Liu, Jingxiang Sun, Tongzheng Ren, Zhuoshu Li, Hao Yang, Yaofeng Sun, Chengqi Deng, Hanwei Xu, Zhenda Xie, and Chong Ruan. 2024.
\newblock \href {https://arxiv.org/abs/2403.05525} {Deepseek-vl: Towards real-world vision-language understanding}.
\newblock \emph{Preprint}, arXiv:2403.05525.

\bibitem[{Ma et~al.(2024)Ma, Zhang, and Zhao}]{ma2024cocoagentcomprehensivecognitivemllm}
Xinbei Ma, Zhuosheng Zhang, and Hai Zhao. 2024.
\newblock \href {https://arxiv.org/abs/2402.11941} {Coco-agent: A comprehensive cognitive mllm agent for smartphone gui automation}.
\newblock \emph{Preprint}, arXiv:2402.11941.

\bibitem[{Nakano et~al.(2022)Nakano, Hilton, Balaji, Wu, Ouyang, Kim, Hesse, Jain, Kosaraju, Saunders, Jiang, Cobbe, Eloundou, Krueger, Button, Knight, Chess, and Schulman}]{nakano2022webgptbrowserassistedquestionansweringhuman}
Reiichiro Nakano, Jacob Hilton, Suchir Balaji, Jeff Wu, Long Ouyang, Christina Kim, Christopher Hesse, Shantanu Jain, Vineet Kosaraju, William Saunders, Xu~Jiang, Karl Cobbe, Tyna Eloundou, Gretchen Krueger, Kevin Button, Matthew Knight, Benjamin Chess, and John Schulman. 2022.
\newblock \href {https://arxiv.org/abs/2112.09332} {Webgpt: Browser-assisted question-answering with human feedback}.
\newblock \emph{Preprint}, arXiv:2112.09332.

\bibitem[{Niu et~al.(2024)Niu, Li, Wang, Fu, Hu, Leng, Kong, Chang, and Wang}]{niu2024screenagentvisionlanguagemodeldriven}
Runliang Niu, Jindong Li, Shiqi Wang, Yali Fu, Xiyu Hu, Xueyuan Leng, He~Kong, Yi~Chang, and Qi~Wang. 2024.
\newblock \href {https://arxiv.org/abs/2402.07945} {Screenagent: A vision language model-driven computer control agent}.
\newblock \emph{Preprint}, arXiv:2402.07945.

\bibitem[{Shen et~al.(2023)Shen, Song, Tan, Li, Lu, and Zhuang}]{shen2023hugginggptsolvingaitasks}
Yongliang Shen, Kaitao Song, Xu~Tan, Dongsheng Li, Weiming Lu, and Yueting Zhuang. 2023.
\newblock \href {https://arxiv.org/abs/2303.17580} {Hugginggpt: Solving ai tasks with chatgpt and its friends in hugging face}.
\newblock \emph{Preprint}, arXiv:2303.17580.

\bibitem[{Team(2024)}]{geminiteam2024gemini15unlockingmultimodal}
Gemini Team. 2024.
\newblock \href {https://arxiv.org/abs/2403.05530} {Gemini 1.5: Unlocking multimodal understanding across millions of tokens of context}.
\newblock \emph{Preprint}, arXiv:2403.05530.

\bibitem[{Wang et~al.(2024{\natexlab{a}})Wang, Xu, Jia, Zhang, Yan, Shen, Zhang, Huang, and Sang}]{wang2024mobileagentv2mobiledeviceoperation}
Junyang Wang, Haiyang Xu, Haitao Jia, Xi~Zhang, Ming Yan, Weizhou Shen, Ji~Zhang, Fei Huang, and Jitao Sang. 2024{\natexlab{a}}.
\newblock \href {https://arxiv.org/abs/2406.01014} {Mobile-agent-v2: Mobile device operation assistant with effective navigation via multi-agent collaboration}.
\newblock \emph{Preprint}, arXiv:2406.01014.

\bibitem[{Wang et~al.(2024{\natexlab{b}})Wang, Xu, Ye, Yan, Shen, Zhang, Huang, and Sang}]{wang2024mobileagentautonomousmultimodalmobile}
Junyang Wang, Haiyang Xu, Jiabo Ye, Ming Yan, Weizhou Shen, Ji~Zhang, Fei Huang, and Jitao Sang. 2024{\natexlab{b}}.
\newblock \href {https://arxiv.org/abs/2401.16158} {Mobile-agent: Autonomous multi-modal mobile device agent with visual perception}.
\newblock \emph{Preprint}, arXiv:2401.16158.

\bibitem[{Wang et~al.(2024{\natexlab{c}})Wang, Bai, Tan, Wang, Fan, Bai, Chen, Liu, Wang, Ge, Fan, Dang, Du, Ren, Men, Liu, Zhou, Zhou, and Lin}]{wang2024qwen2vlenhancingvisionlanguagemodels}
Peng Wang, Shuai Bai, Sinan Tan, Shijie Wang, Zhihao Fan, Jinze Bai, Keqin Chen, Xuejing Liu, Jialin Wang, Wenbin Ge, Yang Fan, Kai Dang, Mengfei Du, Xuancheng Ren, Rui Men, Dayiheng Liu, Chang Zhou, Jingren Zhou, and Junyang Lin. 2024{\natexlab{c}}.
\newblock \href {https://arxiv.org/abs/2409.12191} {Qwen2-vl: Enhancing vision-language model's perception of the world at any resolution}.
\newblock \emph{Preprint}, arXiv:2409.12191.

\bibitem[{Wang et~al.(2025)Wang, Xu, Wang, Zhang, Yan, Zhang, Huang, and Ji}]{wang2025mobileagenteselfevolvingmobileassistant}
Zhenhailong Wang, Haiyang Xu, Junyang Wang, Xi~Zhang, Ming Yan, Ji~Zhang, Fei Huang, and Heng Ji. 2025.
\newblock \href {https://arxiv.org/abs/2501.11733} {Mobile-agent-e: Self-evolving mobile assistant for complex tasks}.
\newblock \emph{Preprint}, arXiv:2501.11733.

\bibitem[{Wolf et~al.(2020)Wolf, Debut, Sanh, Chaumond, Delangue, Moi, Cistac, Rault, Louf, Funtowicz, Davison, Shleifer, von Platen, Ma, Jernite, Plu, Xu, Scao, Gugger, Drame, Lhoest, and Rush}]{wolf2020huggingfacestransformersstateoftheartnatural}
Thomas Wolf, Lysandre Debut, Victor Sanh, Julien Chaumond, Clement Delangue, Anthony Moi, Pierric Cistac, Tim Rault, Rémi Louf, Morgan Funtowicz, Joe Davison, Sam Shleifer, Patrick von Platen, Clara Ma, Yacine Jernite, Julien Plu, Canwen Xu, Teven~Le Scao, Sylvain Gugger, Mariama Drame, Quentin Lhoest, and Alexander~M. Rush. 2020.
\newblock \href {https://arxiv.org/abs/1910.03771} {Huggingface's transformers: State-of-the-art natural language processing}.
\newblock \emph{Preprint}, arXiv:1910.03771.

\bibitem[{Wu et~al.(2024{\natexlab{a}})Wu, Han, Ding, Weng, Liu, Yao, Yu, and Kong}]{wu2024oscopilotgeneralistcomputeragents}
Zhiyong Wu, Chengcheng Han, Zichen Ding, Zhenmin Weng, Zhoumianze Liu, Shunyu Yao, Tao Yu, and Lingpeng Kong. 2024{\natexlab{a}}.
\newblock \href {https://arxiv.org/abs/2402.07456} {Os-copilot: Towards generalist computer agents with self-improvement}.
\newblock \emph{Preprint}, arXiv:2402.07456.

\bibitem[{Wu et~al.(2024{\natexlab{b}})Wu, Wu, Xu, Wang, Sun, Jia, Cheng, Ding, Chen, Liang, and Qiao}]{wu2024osatlasfoundationactionmodel}
Zhiyong Wu, Zhenyu Wu, Fangzhi Xu, Yian Wang, Qiushi Sun, Chengyou Jia, Kanzhi Cheng, Zichen Ding, Liheng Chen, Paul~Pu Liang, and Yu~Qiao. 2024{\natexlab{b}}.
\newblock \href {https://arxiv.org/abs/2410.23218} {Os-atlas: A foundation action model for generalist gui agents}.
\newblock \emph{Preprint}, arXiv:2410.23218.

\bibitem[{Wu et~al.(2024{\natexlab{c}})Wu, Chen, Pan, Liu, Liu, Dai, Gao, Ma, Wu, Wang, Xie, Wu, Hu, Wang, Sun, Li, Piao, Guan, Liu, Xie, You, Dong, Yu, Zhang, Zhao, Wang, and Ruan}]{wu2024deepseekvl2mixtureofexpertsvisionlanguagemodels}
Zhiyu Wu, Xiaokang Chen, Zizheng Pan, Xingchao Liu, Wen Liu, Damai Dai, Huazuo Gao, Yiyang Ma, Chengyue Wu, Bingxuan Wang, Zhenda Xie, Yu~Wu, Kai Hu, Jiawei Wang, Yaofeng Sun, Yukun Li, Yishi Piao, Kang Guan, Aixin Liu, Xin Xie, Yuxiang You, Kai Dong, Xingkai Yu, Haowei Zhang, Liang Zhao, Yisong Wang, and Chong Ruan. 2024{\natexlab{c}}.
\newblock \href {https://arxiv.org/abs/2412.10302} {Deepseek-vl2: Mixture-of-experts vision-language models for advanced multimodal understanding}.
\newblock \emph{Preprint}, arXiv:2412.10302.

\bibitem[{Xu et~al.(2024)Xu, Wang, Wang, Lu, Xie, Saha, Sahoo, Yu, and Xiong}]{xu2024aguvis}
Yiheng Xu, Zekun Wang, Junli Wang, Dunjie Lu, Tianbao Xie, Amrita Saha, Doyen Sahoo, Tao Yu, and Caiming Xiong. 2024.
\newblock \href {https://arxiv.org/abs/2412.04454} {Aguvis: Unified pure vision agents for autonomous gui interaction}.

\bibitem[{Yang et~al.(2023)Yang, Zhang, Li, Zou, Li, and Gao}]{yang2023setofmarkpromptingunleashesextraordinary}
Jianwei Yang, Hao Zhang, Feng Li, Xueyan Zou, Chunyuan Li, and Jianfeng Gao. 2023.
\newblock \href {https://arxiv.org/abs/2310.11441} {Set-of-mark prompting unleashes extraordinary visual grounding in gpt-4v}.
\newblock \emph{Preprint}, arXiv:2310.11441.

\bibitem[{Yang et~al.(2024)Yang, Wang, Li, Luo, Chen, Huang, and Li}]{yang2024ariauivisualgroundinggui}
Yuhao Yang, Yue Wang, Dongxu Li, Ziyang Luo, Bei Chen, Chao Huang, and Junnan Li. 2024.
\newblock \href {https://arxiv.org/abs/2412.16256} {Aria-ui: Visual grounding for gui instructions}.
\newblock \emph{Preprint}, arXiv:2412.16256.

\bibitem[{Ye et~al.(2024)Ye, Xu, Xu, Ye, Yan, Zhou, Wang, Hu, Shi, Shi, Li, Xu, Chen, Tian, Qian, Zhang, Huang, and Zhou}]{ye2024mplugowlmodularizationempowerslarge}
Qinghao Ye, Haiyang Xu, Guohai Xu, Jiabo Ye, Ming Yan, Yiyang Zhou, Junyang Wang, Anwen Hu, Pengcheng Shi, Yaya Shi, Chenliang Li, Yuanhong Xu, Hehong Chen, Junfeng Tian, Qi~Qian, Ji~Zhang, Fei Huang, and Jingren Zhou. 2024.
\newblock \href {https://arxiv.org/abs/2304.14178} {mplug-owl: Modularization empowers large language models with multimodality}.
\newblock \emph{Preprint}, arXiv:2304.14178.

\bibitem[{You et~al.(2024)You, Zhang, Schoop, Weers, Swearngin, Nichols, Yang, and Gan}]{you2024ferretuigroundedmobileui}
Keen You, Haotian Zhang, Eldon Schoop, Floris Weers, Amanda Swearngin, Jeffrey Nichols, Yinfei Yang, and Zhe Gan. 2024.
\newblock \href {https://arxiv.org/abs/2404.05719} {Ferret-ui: Grounded mobile ui understanding with multimodal llms}.
\newblock \emph{Preprint}, arXiv:2404.05719.

\bibitem[{Zhang et~al.(2024)Zhang, Li, He, Zhang, Qiao, Qin, Ma, Kang, Lin, Rajmohan, Zhang, and Zhang}]{zhang2024ufouifocusedagentwindows}
Chaoyun Zhang, Liqun Li, Shilin He, Xu~Zhang, Bo~Qiao, Si~Qin, Minghua Ma, Yu~Kang, Qingwei Lin, Saravan Rajmohan, Dongmei Zhang, and Qi~Zhang. 2024.
\newblock \href {https://arxiv.org/abs/2402.07939} {Ufo: A ui-focused agent for windows os interaction}.
\newblock \emph{Preprint}, arXiv:2402.07939.

\bibitem[{Zhang et~al.(2023)Zhang, Yang, Liu, Han, Chen, Huang, Fu, and Yu}]{zhang2023appagentmultimodalagentssmartphone}
Chi Zhang, Zhao Yang, Jiaxuan Liu, Yucheng Han, Xin Chen, Zebiao Huang, Bin Fu, and Gang Yu. 2023.
\newblock \href {https://arxiv.org/abs/2312.13771} {Appagent: Multimodal agents as smartphone users}.
\newblock \emph{Preprint}, arXiv:2312.13771.

\bibitem[{Zheng et~al.(2024)Zheng, Huang, Xue, Wang, An, and Yan}]{zheng2024agentstudio}
Longtao Zheng, Zhiyuan Huang, Zhenghai Xue, Xinrun Wang, Bo~An, and Shuicheng Yan. 2024.
\newblock Agentstudio: A toolkit for building general virtual agents.
\newblock \emph{arXiv preprint arXiv:2403.17918}.

\end{thebibliography}
